\documentclass{article} 
\usepackage{style,times}

\usepackage{amsmath,amsfonts,bm}

\def\eqref#1{equation~\ref{#1}}

\def\1{\bm{1}}

\DeclareMathAlphabet{\mathsfit}{\encodingdefault}{\sfdefault}{m}{sl}
\SetMathAlphabet{\mathsfit}{bold}{\encodingdefault}{\sfdefault}{bx}{n}

\newcommand{\R}{\mathbb{R}}

\usepackage{url}
\usepackage{titlesec} 
\titlespacing*{\section}
{0pt}{12pt plus 4pt minus 2pt}{6pt plus 2pt minus 2pt} 
\titlespacing*{\subsection}
{0pt}{10pt plus 4pt minus 2pt}{4pt plus 2pt minus 2pt} 

\usepackage{xcolor}
\usepackage{enumitem}
\usepackage{hyperref}
\definecolor{royalblue}{rgb}{0.0, 0.2, 0.7}
\hypersetup{
    colorlinks,
    citecolor=royalblue,
    filecolor=royalblue,
    linkcolor=royalblue,
    urlcolor=royalblue
}
\usepackage{url}
\usepackage{placeins}
\usepackage{todonotes}
\usepackage{tikz}
\setuptodonotes{inline}

\usepackage{booktabs}
\usepackage{graphicx}

\usepackage[font=small,labelfont=bf]{caption}
\usepackage{titlesec}
\titlespacing*{\section}{0pt}{0.0\baselineskip}{0.0\baselineskip}
\newcommand{\mycirc}[1][black]{\Large\textcolor{#1}{\ensuremath\bullet}}
\newcommand{\bftab}{\fontseries{b}\selectfont}
\title{
Towards generalizable single-cell perturbation modeling via conditional Monge Gap

}

\author{
Alice Driessen \thanks{IBM Research Europe, 8803 Rüschlikon, Zurich, Switzerland \\
Department of Biosystems Science and Engineering, ETH Zürich, 4056 Basel, Switzerland} \And
Benedek Harsanyi \thanks{IBM Research Europe, 8803 Rüschlikon, Zurich} \And
Marianna Rapsomaniki \thanks{IBM Research Europe, 8803 Rüschlikon, Zurich Switzerland \\
Biomedical Data Science Center, Lausanne University Hospital and University of Lausanne, Switzerland \\
Swiss Institute of Bioinformatics (SIB)} \And
\centerline{Jannis Born \thanks{IBM Research Europe, 8803 Rüschlikon, Zurich \\ 
Corresponding author}}}

\iclrfinalcopy 
\begin{document}

\maketitle

\begin{abstract}

Learning the response of single-cells to various treatments offers great potential to enable targeted therapies. 
In this context, neural optimal transport (OT) has emerged as a principled methodological framework because it inherently accommodates the challenges of unpaired data induced by cell destruction during data acquisition. 
However, most existing OT approaches are incapable of conditioning on different treatment contexts (e.g., time, drug treatment, drug dosage, or cell type) and we still lack methods that unanimously show promising generalization performance to unseen treatments.
Here, we propose the Conditional Monge Gap which learns OT maps conditionally on arbitrary covariates.
We demonstrate its value in predicting single-cell perturbation responses conditional to one or multiple drugs, a drug dosage, or combinations thereof. 
We find that our conditional models achieve results comparable and sometimes even superior to the condition-specific state-of-the-art on scRNA-seq as well as multiplexed protein imaging data.
Notably, by aggregating data across conditions we perform cross-task learning which unlocks remarkable generalization abilities to unseen drugs or drug dosages, widely outperforming other conditional models in capturing heterogeneity (i.e., higher moments) in the perturbed population.
Finally, by scaling to hundreds of conditions and testing on unseen drugs, we narrow the gap between structure-based and effect-based drug representations, suggesting a promising path to the successful prediction of perturbation effects for unseen treatments.

\end{abstract}

\section{Introduction}
Understanding how cells states change in response to different stimuli is a long-standing question with broad implications for biology and medicine.
Single-cell transcriptomics (scRNA-seq) coupled with high-throughput screening allows capturing the response of heterogeneous cell populations to thousands of genetic~\citep{frangieh2021multimodal} or drug perturbations at once~\citep{dixit2016perturb,ji2021machine,peidli2024scperturb}. 
Such techniques have shown great potential to enable targeted therapies~\citep{ianevski2024single,bai2024single,sinha2024perception}, but as the cost of such experiments remains high, \textit{in silico} modeling of perturbation responses has emerged as an appealing alternative.

While many methods for predicting single-cell perturbation responses have been developed, their vast majority (e.g., scGen~\cite{lotfollahi2019scgen} or CellOT~\cite{bunne2023learning}) is not directly usable for the most impactful clinical application: predicting treatment response for \textit{novel} therapies, such as a new drug, the same drug in a new dosage or another CRISPR-edited T cell cancer immunotherapy.
This necessitates conditional models that can be trained globally across perturbations and can be applied either on known (i.e., in-distribution, ID) or unknown (i.e., out-of-distribution, OOD) perturbations.

Early attempts included different flavors of autoencoders~\citep{lopez2018deep, lotfollahi2019scgen,hetzel2022predicting,lotfollahi2023predicting,yu2022perturbnet,liu2024learning} that predict perturbation effects by solving a linear transformation problem in latent space.
The unconditional model scGen~\citep{lotfollahi2019scgen}
was extended to model combinatorial perturbations and account for covariates such as drug dose and cell types (cf. CPA~\cite{lotfollahi2023predicting}).
Building upon this, chemCPA~\citep{hetzel2022predicting} and PerturbNet~\citep{yu2022perturbnet} are among the few methods to support the prediction of perturbations for unseen drugs.
Concurrently, single-cell foundation models also gained popularity~\citep{cui2024scgpt,yang2022scbert} but in perturbation prediction they have not yet matched linear models~\citep{csendes2024benchmarking,ahlmann2024deep} unless task-specific adapters are used~\citep{maleki2024efficient}.

Ultimately however, all such methods do not capture the underlying cellular heterogeneity observed both within and across samples, as samples could originate from different cell types, conditions, or even patients. 
This is because the main challenge of perturbation modeling stems from the destructive nature of single-cell transcriptomic measurements, implying that cell populations are unpaired.
This motivates the use of optimal transport (OT) to match unperturbed and perturbed populations of cells as probability distributions~\citep{bunne2024optimal}. 
OT has been successfully employed to several similar problems of mapping cellular distributions \citep{klein2024genot}, such as reconstructing cell evolution trajectories \citep{schiebinger_optimal-transport_2019,klein2025mapping}, and aligning single-cell measurements across different omic modalities \citep{gossi_matching_2023,cao2022unified}.

Given two discrete cell distributions, neural OT allows to learn how the unperturbed cells have morphed into the perturbed cells~\citep{bunne2024optimal}. 
By minimizing displacement cost, OT allows to go beyond predicting average effects and better captures higher moments of perturbation effects~\citep{bunne2023learning}, typically measured by maximum mean discrepancy~\citep{borgwardt2006integrating}.
Building upon early OT solvers~\citep{cuturi2013sinkhorn}, more scalable Wasserstein-distance-inspired losses have enabled the training of OT-based generative models~\citep{genevay2018learning,feydy2019interpolating}. 
However, OT methods for single-cell perturbation prediction such as CellOT~\citep{bunne2023learning} or scPRAM~\citep{jiang2024scpram} are typically trained separately (one model per perturbation) due to challenges in learning OT maps conditionally.
The focus for this work is overcoming such challenges and building a lightweight and generic conditional OT method for single-cell perturbation response prediction that generalizes well to unseen drugs.

To learn OT maps with a neural network,  Brenier's theorem~\citeyearpar{brenier1987decomposition} can be leveraged, stating that a unique dual potential exists, which has a gradient equal to the transport map. 
This potential can be represented as a convex function, which gave rise to neural solvers based on input convex neural networks (ICNNs) \citep{amos2017input, makkuva2020optimal}. 
CellOT~\citep{bunne2023learning} is a prominent example building upon these results by leveraging ICNNs. 
However, one drawback of the dual approach is that Brenier's theorem \citeyearpar{brenier1987decomposition} relies on squared Euclidean cost, which is inflexible and unrealistic for high-dimensional data.
Moreover, training ICNNs poses many challenges: it requires special weight initialization, the non-negativity of the weights exacerbates training and the dual training suffers from instability due to its min-max loss function.
While attempts have been made to learn OT maps conditionally via ICNNs through partial ICNNs as in CondOT~\citep{bunne2022supervised}, such networks are even more challenging to learn and have not yet shown compelling results in predicting perturbation effcts for unseen drugs.
To overcome some of these challenges, various novel techniques in OT~\citep{uscidda2023monge,chen2024fast}, quantile regression ~\citep{rosenbergfast,pegoraro2023vector,vedulacontinuous}, flow-matching ~\citep{tong2023improving,pooladian2023multisample} and even quantum computing~\citep{mariella24a,basu2023towards} have been proposed.
Among those, the \textit{Monge Gap}~\citep{uscidda2023monge} stands out due to its simplicity of employing a regularizer to estimate OT maps with any ground cost $c$. The Monge Gap allows to directly parameterize the transport map $T$ and optimizes the debiased version of the primal objective, the Sinkhorn divergence along with the Monge Gap, to ensure $c$-optimality and fit a mapping between the source and the target distribution. 
However, as in CellOT~\citep{bunne2023learning} or scGen~\citep{lotfollahi2019scgen}, the learned maps are local (\textit{i.e.}, unconditional), implying that distinct models are fitted for each condition, which has several shortcomings:
\noindent
\begin{enumerate}[leftmargin=*, label=(\arabic*), topsep=0pt, partopsep=0pt, itemsep=0pt, parsep=0pt, after=\vspace{-0.2\baselineskip}, before=\vspace{-0.5\baselineskip}] 
\item Data for each condition is needed, so no inference can be made for a new condition;
\item The computational cost of training separate models can be significant;
\item There is no inductive bias to accommodate any covariates present in the data;
\item Potential cross-task benefits arising from training concurrently on conditions with similar effects cannot be exploited.
\end{enumerate}

\begin{figure}[t!]
\centering
    \includegraphics[width=1\textwidth]{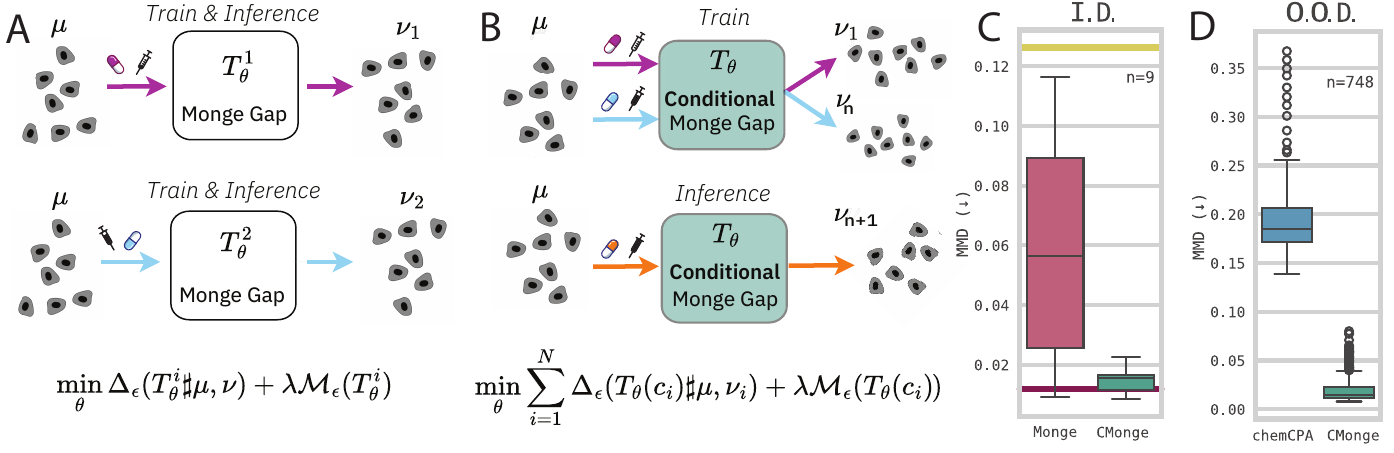}
    \caption{
    \textbf{The Conditional Monge Gap (CMonge) for single-cell perturbation response prediction}. \textbf{A}) Existing methods learn local maps for each perturbation separately. 
    \textbf{B}) We propose to model perturbation responses via a \textit{global} estimator that can be conditioned on a potentially unseen condition $c_i$ at inference time. 
    \textbf{C}) In-distribution results for drug-specific models, with or without dose context information. Boxes show the mean Maximum Mean Discrepancy (MMD) per drug on the SciPlex dataset for the highest dose for 9 drug-specific Monge Gap models and 9 Conditional Monge Gap models with conditional dosage information. 
    Lines indicate the average performance of an identity model (lower bound, yellow) and a single Monge model per condition (36 models, upper bound, red). 
    \textbf{D}) Out-of-distribution results for pan-drug models with drug and dose context information. Boxes show the mean MMD per drug on the SciPlex dataset for the highest dose for CMonge and chemCPA. 
    Both conditional models only require drug structure (SMILES) and are trained and evaluated on all 187 drugs in the SciPlex dataset.}
    \label{fig:overview}
    \vspace*{-0.5pt}
\end{figure}

Here, we propose the Conditional Monge Gap, a novel method to learn a global OT map that can be conditioned on different context variables or covariates (cf.~\autoref{fig:overview}). 
Instead of leveraging partial ICNNs via Brenier's theorem~\citep{bunne2022supervised} or via continuous vector quantile regression~\citep{vedulacontinuous}, we contextualize through the primal objective and directly learn the transportation maps between the source and different target measures. 
To demonstrate that our proposed Conditional Monge Gap is useful for single-cell perturbation prediction, we tested it on different context variables (single and multiple drugs, drug dosage, and combinations thereof), different data splits, and \textit{in} and \textit{out}-of-distribution (ID and OOD) settings. 
We experiment with different condition encodings, including fingerprint-based and effect-driven drug representations.
We find that using a single model for multiple conditions shows little to no performance loss compared to learning a model per condition. 
Indeed, this single model benefits from contextual information and uses cross-task-learning to improve performance. 
Importantly, we show compelling results in predicting perturbation effects for unseen drug therapies -- the Conditional Monge Gap captures well the heterogeneity of cellular response, even for previously unobserved drug therapies.
In direct comparison to another conditional model~\citep{hetzel2022predicting}, the Conditional Monge yields consistently better results in both a small and large dataset setting. 
Together our results show the promise of using the Conditional Monge Gap for single cell perturbation modeling in seen and unseen contexts.

\section{Results}

\subsection{Conditional Optimal Transport Estimators}
For a brief background on optimal transport (OT) see Methods \ref{sec:ot_background}. 
We aim to learn a parametrized OT map $T_{\theta}$ by minimizing the Sinkhorn divergence along with the Monge Gap regularizer~\citep{uscidda2023monge} and $\lambda \geq 0$,
\begin{align}\label{eq:monge}
\min_{\theta} \Delta_{\epsilon}(T_{\theta}\sharp\mu, \nu)+\lambda \mathcal{M}_{\epsilon}(T_\theta),
\end{align}
where $\Delta$ is the differentiable Sinkhorn divergence specified in Methods \ref{sec:ot_background} \citep{ramdas_wasserstein_2017, feydy2019interpolating, salimans_improving_2018, genevay2018learning}, $\epsilon$ is the strength of entropic regularization used for solving $\Delta$ and $\mathcal{M}$. $T_\theta \sharp\mu$ is the push-forward of measure $\mu$, which should approximate the target measure $\nu$. $\lambda$ is the regularization strength for the Monge Gap regularizer $\mathcal{M}_{\epsilon}$. $\mathcal{M}_{\epsilon}$ is the Monge Gap which ensures cost optimality w.r.t. the squared Euclidean distance $c(x,y)=\lvert \lvert x-y \rvert \rvert^2$ between the source and transported samples and can be estimated from samples, by
\begin{align}
    \mathcal{M}_{\epsilon}(T_{\theta})=\frac{1}{n}\sum_{i=1}^n c(x_i, T_{\theta}(x_i))-W_{\epsilon}(\mu, T_{\theta}\sharp\mu).
\end{align}
Where $x_i$ are samples from the source distribution $\mu$. However, in a conditional setting, multiple target probability measures are labeled with a context $c_i$ (note the difference from the cost $c$), such that $(c_i, \mu, \nu_i) \in  \mathcal{P}(\mathbb{R}^k\times\mathbb{R}^d)\times\mathcal{P}(\mathbb{R}^d)$ for $i\in\{1,\ldots K\}$. 
Instead of learning distinct mappings $T_i\sharp(\mu)\approx\nu_i$ we build a global parametrization $T_{\theta}: \mathbb{R}^k\times\mathbb{R}^d\to \mathbb{R}^d$ that captures information contextually. 
Our proposed loss extends~\autoref{eq:monge}, by simultaneously optimizing for each of the $K$ conditions 
\begin{align}\label{eq:cond_monge}
 \min_{\theta} \sum_{i=1}^K\Delta_{\epsilon}(T_{\theta}(c_i)\sharp\mu, \nu_i)+\lambda \mathcal{M}_{\epsilon}(T_\theta(c_i)).
\end{align}
\subsection{Datasets}
\label{sec:31}
We used two datasets for training and evaluating our models. First, the SciPlex3 dataset~\citep{srivatsan2020massively} contains single-cell profiles from three human cancer cell lines (namely A549, K562, and MCF7) that were exposed to a total of $188$ compounds. 187 drugs that were administered at four different doses ($10$ nM, $100$ nM, $1000$ nM, and $10000$ nM) and a control solution. We performed experiments on a selection of 9 different drugs as in ~\citet{uscidda2023monge} and on the full range of compounds, similar to \citet{hetzel2022predicting}. 

We used the preprocessed data from \citet{lotfollahi2019scgen}, which has been preprocessed with library size normalization, cell and gene filtering, and \texttt{log1p} transformation. The dataset consists of $762,039$ single-cell measurements, out of which $17,565$ belong to the control population, and on average $4,032$ observations to each drug and drug dosage condition. During training and evaluation, we only consider the $1000$ highly variable genes (HVG), computed by \citet{lotfollahi2019scgen}. As many genes are left unaffected, instead of evaluating in the $1000$-dimensional gene space, we restrict ourselves to only the top $50$ differentially expressed marker genes obtained through gene ranking~\citep{wolf2018scanpy}.

Secondly, the 4i dataset contains cellular and nuclear measurements of $97,748$ cells ($10,995$ controls) of two lines of melanoma tumors treated with one of $35$ cancer therapies, each involving $\sim2,500$ cells. We obtained the preprocessed data from~\cite{bunne2022supervised}, resulting in $48$ features regarding cellular marker expression and cell shape. 

\subsection{Architecture}
\texttt{CMonge} is trained in two steps. First, we reduced the dimensionality of the $1000$-dimensional gene expression to facilitate OT-map learning. Following \citet{bunne2023learning}, we encoded gene expression data into a $k=50$-dimensional latent space by training a vanilla autoencoder with an encoder $E_{\phi}:\R^{d} \to \R^{k}$ and a decoder $D_{\theta}:\R^{k} \to \R^{d}$. Both $E_{\phi}$ and $D_{\theta}$ are parameterized by multi-layer perceptrons (MLP). The entire autoencoder is optimized using a mean-squared error reconstruction loss.
In the second phase, an OT map is learned between the encoded unperturbed and perturbed cells by optimizing~\autoref{eq:cond_monge}. During this phase, we used the same training set as for the autoencoder, and the autoencoder weights were frozen. The map is parameterized by $v_i=T_\varphi(c_i, z_i)$, where $z_i\in\mathbb{R}^{k}$ is the encoded gene expression, and $c_i$ is the context vector. Throughout this paper, we refer to $T_\varphi$ as the Monge network. Predictions for the cell state space are made by shifting with the learned perturbation and decoding the result $D_{\theta}(E_{\phi}(x_i)+v_i)$. The next section describes how contextual information can be encoded and incorporated into the architecture.

\subsection{Encoding the condition}
We evaluated \texttt{CMonge} in different conditional settings, namely for drug, dosage, and the combination of drug and dosage (DrugDose). The dosage is encoded by transformation of $\text{dose}\to \log{(\text{dose})}$. We considered two strategies for the drug encoding, namely RDkit, and a mode-of-action (MoA) embedding. RDKit is a fingerprint-based molecular representation of 194 features including atom-based and bond-based features including atom type, number of bonds, formal charge, atom mass, and number of hydrogen atoms (for the full list see \citet{yang2019analyzing}). The RDkit embedding is extracted from the SMILES representation of the underlying drug, following \citet{lotfollahi2023predicting}. The MoA embedding is a data-driven approach based on multidimensional scaling (MDS) embeddings, generated by calculating pairwise Wasserstein distances between individual target populations, following \citet{bunne2022supervised}. This approach ensures that perturbations with similar effects in the feature spance are represented closely within the embedding. We calculated a 10-dimensional MDS embedding by employing the majorization algorithm \texttt{SMACOF}~\citep{de2005applications} to minimize stress.

Following \citet{lotfollahi2023predicting}, we consider an initial drug embedding based on RDkit or MoA $h_{i}\in \mathbb{R}^m$, and the transformed dosage $s_i\in\mathbb{R}$. The drug embedding $h_i$ is passed to a drug encoder $W_{\text{drug}}:\mathbb{R}^{m}\to \mathbb{R}^{\varphi_0}$, and the concatenation of $h_i$ and $s_i$ to a drug-dosage encoder $W_{\text{dose}}:\mathbb{R}^{m+1}\to \mathbb{R}$. The final conditional embedding is obtained by the concatenation of the drug and drug-dosage encodings:
\begin{align}
    W_{\text{drug}}(h_i)=z_{i}^{\text{drug}},~~~ W_{\text{dose}}(h_i, s_i)=z_i^{\text{dose}}, ~~~  c_i=(z_{i}^{\text{drug}},z_i^{\text{dose}}).
\end{align}
In the case of combinatorial treatments, that is, multiple drugs in the same condition, we applied a DeepSets layer with average pooling \citep{zaheer2017deep}. An initial embedding $h_i$ for each drug in the combination is obtained and passed to the drug encoder $W_{drug}$. The $W_{drug}$ is the same for all drugs in the combination. The embeddings $W_{drug}=z_i^{single drug}$ per drug are averaged to obtain the final embedding $z_i^{drug}$. The resulting $z_i^{drug}$ has the same dimensions as if one drug is embedded. 

Our context $c_i\in \mathbb{R}^{50}\times \mathbb{R}$, is concatenated with the unperturbed single-cell observation $(z_i)$ and passed to the Monge network $T_\varphi$. $T_\varphi$ is an MLP with sizes $\boldsymbol{\varphi}$ and Gaussian Error Linear Units (GELU) \citep{hendrycks2016gaussian} activation functions. $T_\varphi$ learns the effect of the perturbation, such that the final prediction is the addition of the unperturbed single-cell observation: $z_i + T_\varphi(c_i, z_i)$

\subsection{Evaluation settings}
\label{subsec:eval_settings}
We evaluated with the $R^2$ metric between the perturbed and predicted feature means, the Wasserstein distance (Methods \autoref{eq:kantor}), and the Maximum Mean Discrepancy (MMD). 
For the SciPlex dataset, we trained and evaluated the Conditional Monge Gap in the following scenarios:
\setlist{nolistsep}
\begin{enumerate}[leftmargin=*,noitemsep]
    \item \texttt{Monge}: As a hypothetical upper bound on performance, we fit separate Monge Gap models, one per drug-dosage pair. These models do not have any context (cf.~\citet{uscidda2023monge}).
    \item \texttt{Monge-Drug} / \texttt{Monge-DrugDose}: To motivate the contextual settings, we fit a Monge Gap model that is trained on multiple conditions but unaware of conditional information and evaluate it on conditions seen during training. 
    \begin{itemize}
        \item Monge-Dose-ID: a homogeneous model for each drug, using data from all four dosages. 
        \item Monge-DrugDose-ID: one model on all conditions (all drug-dosage pairs). 
        \item Monge-Dose-OOD: a model for each drug and left out different dosages during training. 
        \item Monge-DrugDose-OOD: a model trained on all conditions but the dosages of the held-out drug(s). 
    \end{itemize} 
    \item \texttt{CMonge-Dose-ID} / \texttt{CMonge-Dose-OOD}: We fit conditional models for each drug with the scalar dose as context. The ID setting sees all dosages during training. For the OOD setting, we left out different dosages during training, thus creating interpolation and extrapolation settings. 
    \item \texttt{CMonge-DrugDose-RDKit / CMonge-DrugDose-MoA}: A single model fitted to all data, conditioned on drug and dosage context. To encode the drug, we compare fingerprints (\texttt{RDKit}) to a data-driven approach (\texttt{MoA}).
    \begin{itemize}
        \item CMonge-DrugDose-x-ID: All conditions are seen during training
        \item CMonge-DrugDose-x-OOD: All dosages of one or more drugs are held during training for evaluation. 
    \end{itemize}      
\end{enumerate}

\subsection{Conditional information improves in distribution prediction}
\begin{figure}[!htb]
    \centering
    \includegraphics[width=1\textwidth]{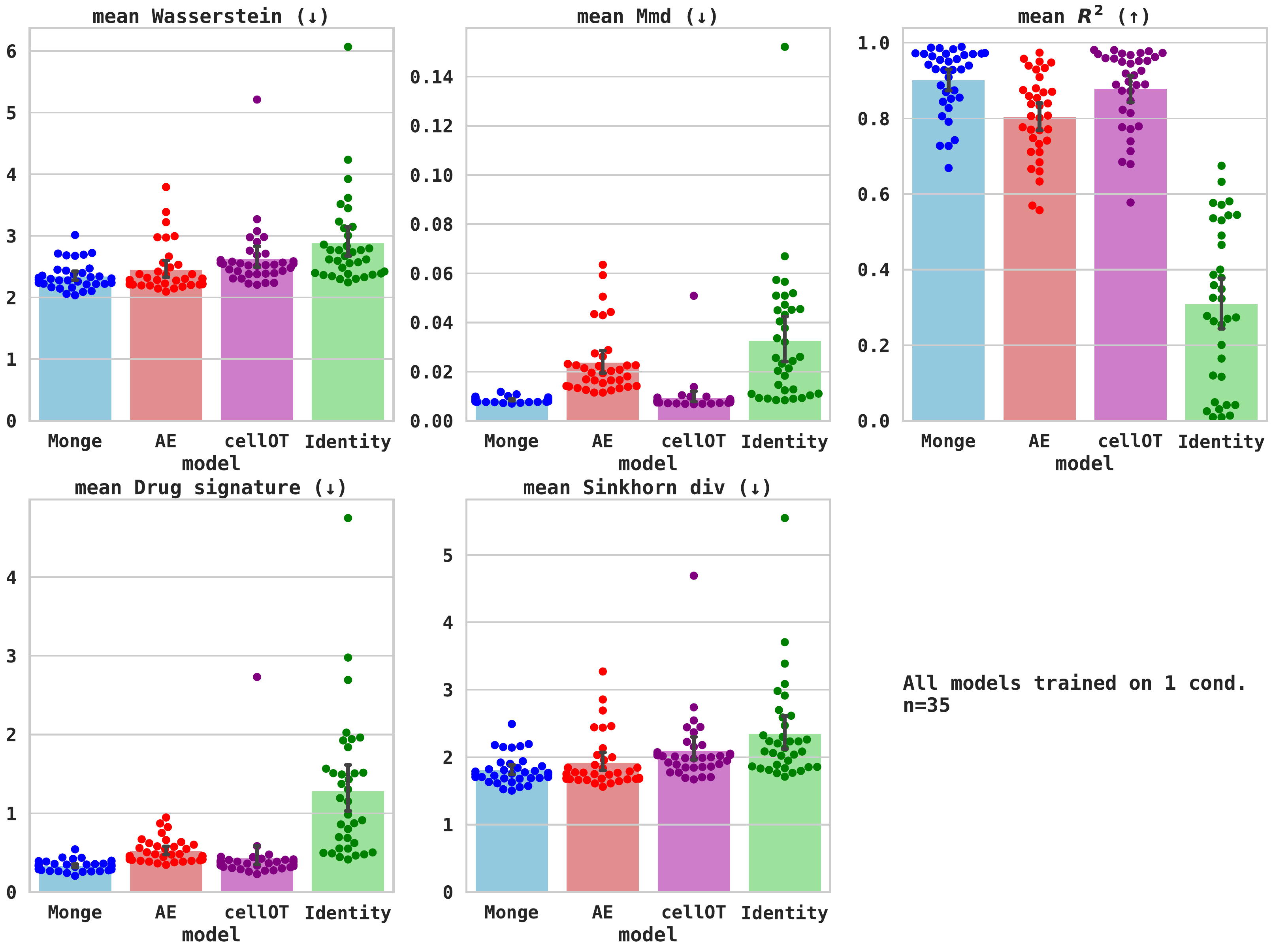}
    \caption{Evaluation of perturbation prediction on the 4i dataset. Each point corresponds to a model trained on one of 35 treatments.}
\label{fig:4i_bar}
\end{figure}

In our experiments, we seek to assess whether adding contextual information enhances the performance of inferring single-cell perturbation responses. Among the plethora of perturbation modeling techniques, we first verified our choice of extending the Monge Gap by comparing it against an ICNN (as in CellOT, \cite{bunne2023learning}) and an autoencoder in an unconditional setting on the 4i data. This experiment revealed the superiority of the Monge Gap (cf. \autoref{fig:4i_bar} and ~\autoref{tab:4i_means_std}). 

\paragraph{SciPlex} We then trained and evaluated our novel conditional Monge map technique on the SciPlex data for the settings mentioned in Methods \ref{subsec:eval_settings}. We compared the trained models to an identity setting, where no optimal transport is used but the source distribution is directly taken as the prediction for the target distribution. 
The high performance of the identity baseline per condition (drug-dose combination) in \autoref{table:dose_ID} shows that the average drug effect is weak on the DEGs over the lower dosages. For the highest dosage of $10000$ nM, the drugs have a significantly stronger effect, making it harder to learn the perturbation responses. 
The difficulty of learning to generalize to the highest dosage is also evidenced by the UMAPs in~\autoref{fig:umap}. For all nine drugs, cells from the three cell types form distinct clusters and for a majority of drugs, the cells belonging to the $10,000$nM condition sit on the edge of the cell-type clusters, sometimes even forming their own clusters. 
The 36 condition-specific Monge models obtain good results across all drugs and dosages and represent an upper bound of the performance. 

We started with only the dosage as context. In a setting where no conditional model is available, one could use a Monge model to predict data for various dosages without supplying conditional information. These Monge-Dose-ID models show a significant performance drop with respect to the Monge upper bound, as the best Monge-Dose-ID can learn is the average response per drug over all dosages. This performance drop motivates us to introduce the CMonge-Dose-ID, which leverages dosage information. CMonge-Dose-ID achieves similar results compared to the upper bound (\textit{i.e.}, the 36 specific Monge models), recovers most of the performance loss of Monge-Dose-ID,  and outperforms Monge-Dose-ID for all dosages. Note that 9 CMonge-Dose-ID models perform on par with 36 condition-specific Monge models, even for the most difficult setting at a dosage of $10000$ nM despite using four times fewer models. 

\tabcolsep=0.10cm
\begin{table}[!htb]
\centering\footnotesize
\caption{Evaluation of conditional and unconditional dose experiments in the ID and OOD setting. Results are compared based on the Coefficient of Determination ($R^2$) between the predicted and target feature means. The average and standard deviation are reported for the 9 experiments/drugs per dosage. "Conditions seen" refers to how many of the 36 drug-dose conditions were in the training set per model.}
\label{table:dose_ID}
\begin{tabular}{lccccccccc}
    \toprule
    Model           & Conditions& \multicolumn{2}{ c }{Context}  & \multicolumn{4}{ c }{Dosage (nM)}\\
                    & seen      &  Drug & Dose                   & 10 & 100 & 1000 & 10000    \\
    \midrule 
    \midrule
    Identity        &           & &            & $0.748_{0.127}$           & $0.655_{0.261}$           & $0.504_{0.332}$           & $0.227_{0.212}$           \\
    Monge           &  1        & &            & $\mathbf{0.950_{0.020}}$  & $\mathbf{0.935_{0.042}}$  & $\mathbf{0.960_{0.025}}$  & $\mathbf{0.978_{0.029}}$  \\
    Monge-Dose-ID   &  4        & &            & $0.750_{0.254}$           & $0.767_{0.231}$           & $0.885_{0.099}$           & $0.694_{0.272}$           \\
    CMonge-Dose-ID  &  4        & & \checkmark & $ 0.882_{0.185}$          & $0.905_{0.124}$           & $0.905_{0.093}$           & $0.974_{0.026}$           \\
    \midrule 
    \midrule
    Monge-Dose-OOD  &  3        & &            & $0.614_{0.349}$           & $0.726_{0.219}$           & $0.856_{0.112}$           & $0.322_{0.348}$           \\
    CMonge-Dose-OOD &  3        & & \checkmark & $0.864_{0.197}$  & $0.931_{0.058}$  & $0.878_{0.118}$           & $0.527_{0.311}$           \\
    \bottomrule
    \end{tabular}
\end{table}

\begin{figure}[!htb]
    \centering
    \includegraphics[width=1\textwidth]{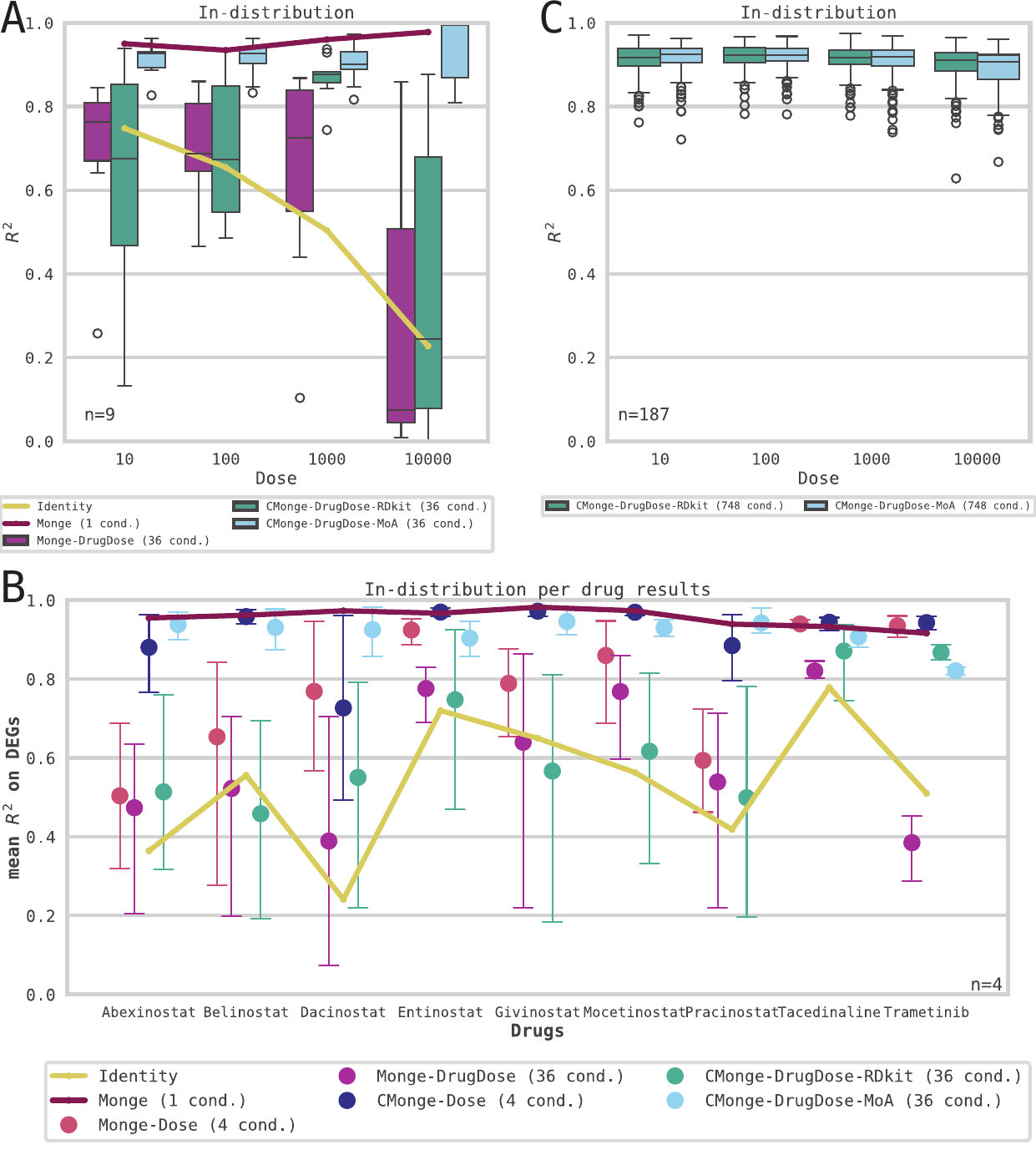}
    \caption{Comparison of the different conditional and unconditional Monge methods in the ID setting using the $R^2$ metric. For each model it is indicated how many of the 36 drug-dose conditions were in the training set per individual model. A) In-distribution results on 9 selected drugs where CMonge is conditioned on drug and dose.
    See \autoref{table:drug_dose_id} for mean and standard deviations.
    B) Results are grouped by drug and show the $R^2$ metric for both the dose and drug+dose conditioning. Each point represents the mean performance of the model over the four dosages. Error bars represent the 95\% confidence interval. Note that CMonge-Dose and Monge-Dose are models per-drug (trained on only one drug), whereas CMonge-DrugDose are pan-condition models where one model is trained on all drugs and dosages. Monge is one model per condition (36 models in total). 
    C) In-distribution results on all drugs in the SciPlex dataset where CMonge is conditioned on drug and dose. See \autoref{table:large_drug_dose_id} for mean and standard deviations.}
    \label{fig:sciplex_drugdose_ID}
\end{figure}

We next sought to investigate the more challenging setting of learning a global map contextualized on both drug and dosage. Therefore, we included a drug embedding based on gene expression similarity (Mode of Action or MoA) or a molecular fingerprint (RDkit) and trained a single model on all 36 conditions. \autoref{fig:sciplex_drugdose_ID}A summarizes the results for CMonge-DrugDose-RDkit-ID and CMonge-DrugDose-MoA-ID. Again we used the identity as a lower bound, 36 condition-specific models as the upper bound and an unconditional Monge-DrugDose-ID as naive approach. The unconditional Monge model shows similar performance as the identity baseline, indicating that a condition-unaware model is not powerful enough to learn the perturbation patterns. 
The best-performing model leverages the MoA embedding (CMonge-DrugDose-MoA-ID).
It clearly outperforms the unconditional model and the CMonge-DrugDose-RDkit-ID model. Notably, the CMonge-DrugDose-MoA-ID is on par with our upper bound setting of 36 condition-specific models, despite only using a single model of all conditions. 
Additionally, including conditional drug information improves prediction over only including dosage information as CMonge-DrugDose-MoA-ID outperforms CMonge-Dose-ID for the lower two dosages (\autoref{table:dose_ID} compared to \autoref{table:drug_dose_id}). The model leveraging the RDkit embedding shows little improvement over the identity and unconditional setting in the two lower dosages. 
However, at the two higher dosages, notably $1000$ nM, the conditional RDkit model shows better $R^2$ scores than the two baselines (\autoref{table:drug_dose_id}). 

These results indicate that we can replace condition-specific models with a single conditional one. To further assess this ambitious question,~\autoref{fig:line36} aggregates performance across drugs and compares the 36 individual ICNN and Monge models to the nine, drug-specific CMonge-Dose-ID models as well as the single CMonge-DrugDose-MoA-ID model. Albeit the unfair comparison of a single model to 36 individual and unconditional models of identical size, the CMonge-DrugDose-MoA-ID achieves a good, yet overall inferior $R^2$ in capturing DEG feature means. It seems that the  CMonge models were predominantly driven by the highest dosage (cf.~\autoref{fig:line36}), which induced the strongest perturbation effect (cf.~\autoref{fig:umap}) but could potentially be mitigated with more careful training. Importantly, however, the CMonge-DrugDose-MoA-ID model captures \textit{better} the higher moments of the distribution than the $36$ condition-specific models (as measured by the Wasserstein distance,~\autoref{fig:line36} \textit{right}). This is a critical finding that underlines the advantages of our method and is further supported by the numerical performances (\autoref{table:main_w}) and the barplot (cf.~\autoref{fig:point_w}) of the Wasserstein distance.

In~\autoref{fig:sciplex_drugdose_ID}B, it can be seen that the performance of the RDkit embedding varies highly among drugs whereas the MoA embeddings consistently yield performance very close to the upper bound (i.e., the 36 condition-specific Monge models). 
However, the MoA, unlike the RDKit features, requires the availability of a small population of perturbed cells to compute the embedding.  
Likely, the $194$ dimensional RDKit-fingerprint introduces too much noise compared to the $50$ dimensional MOA signal, resulting in poor performance for dosages that cause little difference. 
We thus suspect that more drugs are needed to learn how to leverage molecular fingerprints for conditioning the optimal transport map.

To verify if the CMonge models using the RDkit embedding improve with an increasing number of conditions, we trained CMonge on all drugs present in the SciPlex dataset encompassing 187 drugs and a total of 748 conditions. We also increased the the optimal transport map size and the embedding size for the features, such that the embedding size of the features and the RDkit embedding is more comparable (see Methods \ref{sec:app_model_size} for model sizes). \autoref{fig:sciplex_drugdose_ID}C shows the results of the larger CMonge models on all drugs in an ID setting. Both models achieve good performance in this settings with average $R^2$ values over 0.8 and $R^2$-values over 0.6 for all drugs. Strikingly, the RDkit-based CMonge benefits clearly from the additional parameters and conditions as it now performs on par with the MoA-based CMonge, which is something we did not observe when training the smaller models on only nine drugs. 

\begin{figure}[!htb]
    \centering
    \includegraphics[width=1\textwidth]{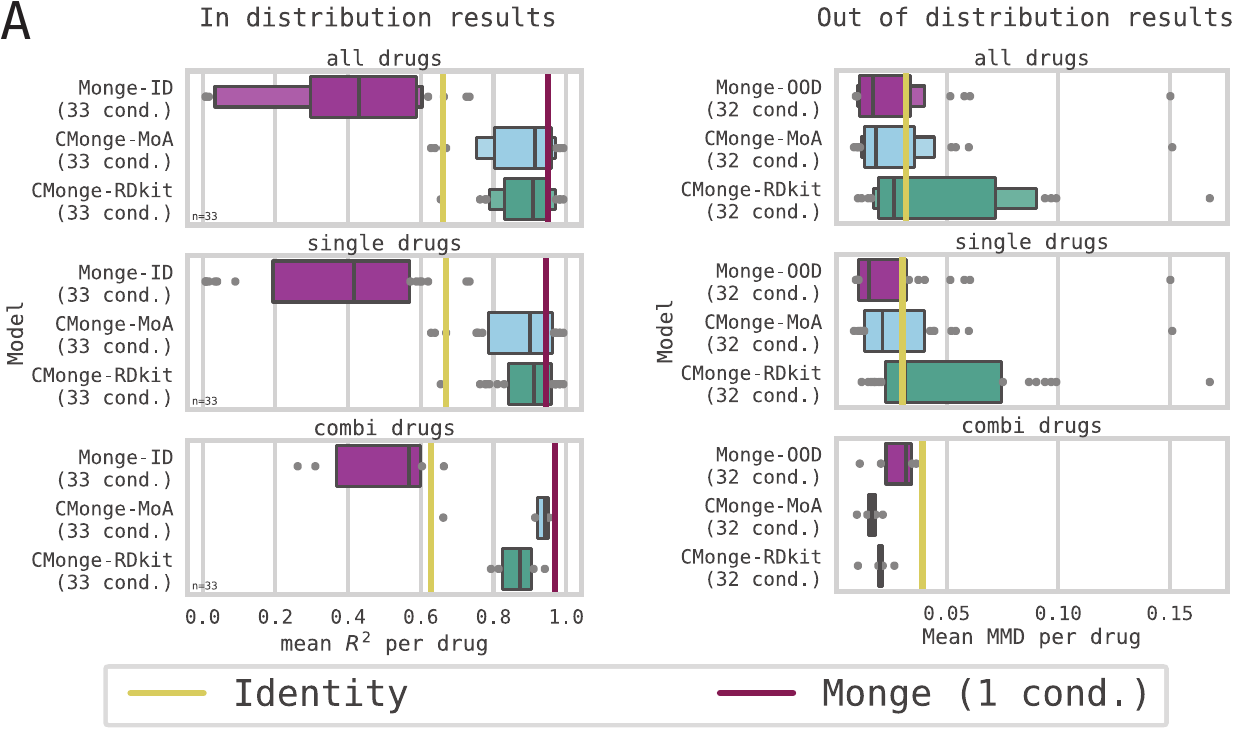}
    \caption{
    Comparison of different Monge, Conditional Monge (CMonge), and the Identity models using the mean $R^2$ or Maximum Mean Discrepancy (MMD). Panels show overall performance, performance for single treatments, and combinatorial treatments. For each model it is indicated how many of the 33 treatments were in the training set per individual model.
    A) In distribution results showing $R^2$. The Monge model is trained as one model that sees all conditions, but is not aware of conditional information. The CMonge models are likewise trained on all conditions, but do get conditional information with the Mode-of-Action (MoA) or RDkit embedding.  
    B) Out of distribution (OOD) results showing MMD. All models are trained in a leave-one-drug-out setting. This means that each boxplot shows the result over 33 drugs, from 33 models. The Monge model is also trained in a leave-one-drug-out setting, but does not incorporate conditional information.}
    \label{fig:4i_R2}
\end{figure}

\paragraph{4i} The 4i dataset contains 35 treatments, consisting of 27 single drug treatments and eight treatments of two or more drugs. We excluded two combinatorial treatments because they did not occur as a single treatment, since we based the MoA drug embedding on the single treatments (see Methods \ref{sec:cond_4i_treatments} for details). This leaves us with 33 single treatments and six combinatorial treatments. \autoref{fig:4i_R2}A shows that also for the 4i dataset adding conditional information improves predictive performance for the $R^2$ metric compared to the identity and a Monge model unaware of conditional information (see Appendix \autoref{fig:4i_EMD} for the Wasserstein distance). For this dataset, the difference between the RDkit- and MoA-based models is smaller than we observed for the nine drugs in SciPlex dataset. Again indicting that more drugs helps in learning how to leverage the RDkit embedding.

Together, these experiments show the benefits of including conditional information in Monge Gap models. Conditional information allows for cross-task learning and reduces compute power with significant performance loss. This suggests that conditional Monge models are generally preferable to unconditional models. 

\subsection{Conditional information allows for out-of-distribution prediction}
Including conditional information has the compelling benefit of allowing predictions for unseen conditions. Therefore, we next investigated how CMonge models generalize to unseen conditions on the 4i dataset and the SciPlex dataset. 
The OOD setting we no longer have an upper-bound baseline model as a Monge model per condition does not yield a model for an unseen condition.

\paragraph{4i OOD} We ran 33 CMonge-MoA and 33 CMonge-RDkit experiments, leaving one treatment out for the 4i dataset. We compared the CMonge models again to the identity baseline and an unconditional Monge baseline. We observed that the overall perturbational effects in this dataset are small: the identity baselines achieves an average $R^2$ of more than 0.6, indicating that the first moment of the distribution (feature means) remains close to the control population. 
Therefore we investigated the performance of CMonge using the Maximum Mean Discrepancy (MMD), as this metrics captures higher moments of the distribution, The CMonge-MoA outperforms the identity for both single drugs and the combination therapies in terms  MMD (\autoref{fig:4i_R2}B, see Appendix \autoref{fig:4i_EMD} for Wasserstein distance). 
For combinatorial therapies, the inclusion of conditional information allows to widely outperform the global, unconditional Monge model.

The small effect sizes, as evidenced by a good identity performance, together with the relatively low amount of $33$ drugs, minimize the improvements than can be gained by introducing conditional information which explains the minimal differences between the baselines and the CMonge models for single drugs.

\paragraph{SciPlex OOD} For the Sciplex dataset we first tested the generalization to unseen dosages by training drug-specific models but holding one of the four dosages out. Not using conditional information in the OOD setting leads to poor performance similar to the identity baseline, as seen in \autoref{table:dose_ID} for Monge-Dose-OOD. OOD prediction clearly improves when conditional dose information is used, as the CMonge-Dose-OOD outperforms the unconditional counterpart and the identity baseline for all dosages. The CMonge-Dose-OOD performs even better than or equal to a conditional model trained on all four dosages (CMonge-Dose-ID), except for the highest dosage. As mentioned above, the highest dosage causes the strongest perturbation effect and is the most difficult setting. Although the generalization to this highest dosage is indeed hard, the CMonge-Dose-OOD outperforms the identity and unconditional settings for this dosage. This indicates that including the conditional information improves the OOD prediction. 

\begin{figure}[!htb]
    \centering
    \includegraphics[width=1\textwidth]{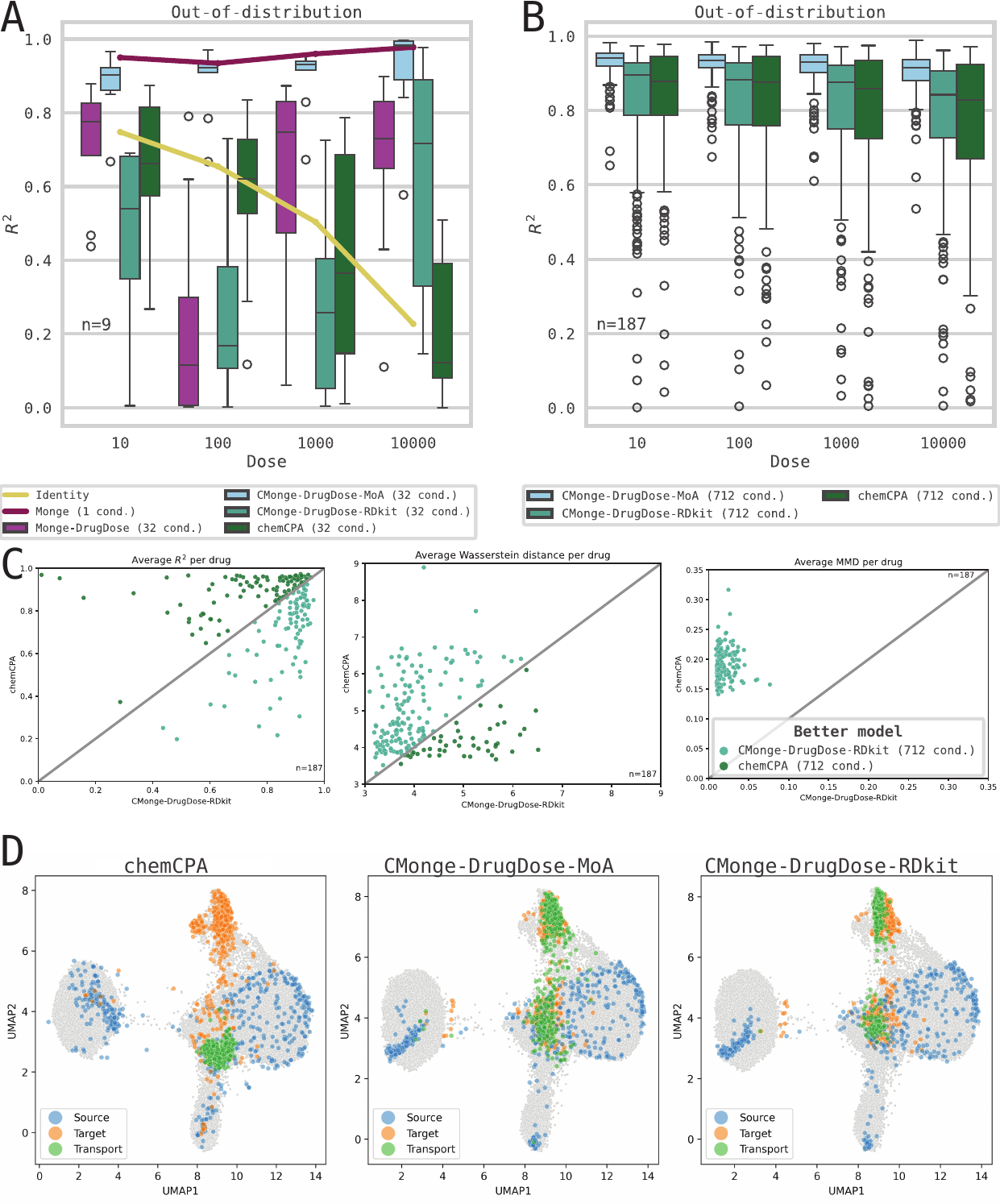}
    \caption{Out-of distribution results on the SciPlex dataset for dose and drug-dose contexts. For each model it is indicated how many of the 36 or 748 drug-dose conditions were in the training set per individual model. 
    A) $R^2$ metric for the dose OOD setting on the selected 9 drugs. Horizontal lines indicate upper and lower bounds of performance. Results are split by dose and shown is the distribution over 9 drugs.
    B) OOD results for all drugs in the SciPlex dataset. Boxplots show the $R^2$ per drug, shown per dosage.
    C) Direct comparison for chemCPA and the CMonge-DrugDose-RDkit for the $R^2$, the Wasserstein distance and the MMD. Each dot is the average performance for a drug over all dosages. Points are colored by wether CMonge or RDkit achieves better performance.
    D) UMAPs for the OOD setting 'Abexinostat-10000' for models trained on the selected 9 drugs. Source, target and transport are taken from one training batch (n=512) and the grey background is the UMAP obtained from all 36 conditions (see Appendix \autoref{fig:extra_umaps}).}
    \label{fig:sciplex_ood}
\end{figure}

Next, we investigated whether predictions for unseen drugs also benefit from conditional information. We performed nine CMonge experiments, always leaving the four dosages of one drug out of the training set for evaluation. We compared CMonge to chemCPA, a SOTA approach that allows perturbation response prediction for unseen drugs. The results in \autoref{fig:sciplex_ood}A, Appendix \autoref{table:drug_dose_ood} and \autoref{table:main_w} reveal that CMonge-MoA widely outperforms chemCPA in all settings. A compelling finding is that CMonge-MoA can almost match its hypothetical upper bound (cf. purple line~\autoref{fig:sciplex_ood}A), i.e., given an unseen drug it yields predictions that are comparable to those obtained after \textit{training} a condition-specific Monge model. This applies to both first moments ($R^2$) and higher moments (Wasserstein distance) of the cell distributions. Additional evidence on this can be observed in Appendix \autoref{fig:ood_sciplex_pointplot_R2}. 

Only for the drug trametinib, CMonge-MoA-OOD performs worse than the drug-specific Dose-OOD models, as do all other DrugDose models. Interestingly, trametinib is the only drug out of the nine drugs that does not affect epigenetic regulation but affects tyrosine kinase signaling (\cite{srivatsan2020massively}). Its mode of action is thus distinctly different from the other drugs, which complicates an OOD prediction. Additionally, we observed higher Wasserstein distances for mocetinostat, which can be explained by a larger distance between the MoA embedding of mocetinostat conditions and the other conditions (\autoref{fig:point_w_ood} \& \autoref{fig:MoA_embed_dist_box}). This higher distance means that the model needs to extrapolate further from what it has seen during training, also making this a particularly difficult drug to predict.

Regarding the remaining OOD results in~\autoref{fig:sciplex_ood}A, it can be seen that CMonge-RDkit outperforms chemCPA for the most difficult setting (highest dose).
However, similar to the ID setting, the CMonge-RDkit struggles to perform well on all dosages which we attribute to the low number of training drugs (8) and the few parameters in the CMonge model (23K). Moreover, beyond the RDKit drug embedding, chemCPA additionally leverages cell line information, and also uses more parameters than CMonge. 
%
We therefore also investigated for the OOD setting whether increasing model size and the number of conditions improves performance for CMonge-RDkit. 
To that end, we performed 21-fold cross-validation by leaving 5\% of drugs (9 drugs) out and training on the remaining conditions. The RDkit-based model gains much performance from increased data and model size, reaching performance close to the MoA-based models. Notably, in this setting CMonge-RDkit outperforms chemCPA for all metrics, especially for the highest dose (\autoref{fig:sciplex_ood}B,C, \autoref{table:large_drug_dose_ood_high_dose}). This difference is more evident for metrics that capture higher moments of the distribution such as the Wasserstein distance and the MMD. CMonge-RDkit maintains high performance over all dosages (see \autoref{table:large_drug_dose_ood} \& \ref{table:large_drug_dose_ood_wasserstein}). 

We also visualized the predictions of CMonge-MoA, CMonge-RDkit and chemCPA on UMAPs showing the source, target, and transport of a single batch for one condition (\autoref{fig:sciplex_ood}D). 
The transport of the CMonge models clearly mixes with the target distribution and shows a good mixture. However, the chemCPA model predictions seem to capture the average but not the full distribution of cells (see Appendix \autoref{fig:extra_umaps} for other drugs). A limitation of chemCPA is that it predicts a mean and variance per cell and gene, thus the Wasserstein distance or MMD cannot be calculated directly. Therefore, to go from chemCPA predictions to synthetic cells, we used the predicted mean as the predicted gene expression of a cell. Although other options are possible, some sort of manipulation of predictions is always necessary to go from chemCPA predictions to counterfactual predictions. Instead, CMonge readily predicts single-cell gene expression values, accounting for stochasticity. 
This final, larger-scale OOD experiment revealed that, with an appropriate number of conditions, the gap to a perturbation-based context like the MoA can be substantially narrowed by utilizing drug structural information alone (see \autoref{table:large_drug_dose_ood} \& \ref{table:large_drug_dose_ood_wasserstein} for all dosages).  

\tabcolsep=0.10cm
\begin{table}[!htb]
\centering\footnotesize
\caption{Evaluation of conditional drug and dose experiments for the highest dose of 10000nM. Results are compared based on the Coefficient of Determination between the predicted and target feature means ($R^2$) and the Wasserstein distance between predicted and measured cells. Results show average and standard deviation across all drugs, evaluated in a leave-9-drugs-out cross-validation setting. DD: DrugDose. "Conditions seen" indicates how many of the 748 drug-dose conditions were in the training set per model.}
\label{table:large_drug_dose_ood_high_dose}
\begin{tabular}{lccccccc}
    \toprule
    Model           & Conditions& \multicolumn{2}{ c }{Context}  &   \multicolumn{3}{ c }{Metric}\\
                    & seen      & Drug & Dose                    & $R^2$ ($\uparrow$) & Wasserstein distance ($\downarrow$) & MMD ($\downarrow$) \\
    \midrule 
    \midrule
    CMonge-DD-MoA-OOD      &  712 & \checkmark & \checkmark & $\mathbf{0.900_{0.059}}$   & $\mathbf{3.873_{0.643}}$ &  $\mathbf{0.013_{0.005}}$  \\
    CMonge-DD-RDkit-OOD    &  712 & \checkmark & \checkmark & $0.781_{0.187}$            & $4.232_{0.896}$  & $0.020_{0.012}$            \\
    chemCPA-DDCellLine-OOD &  712 & \checkmark & \checkmark & $0.760_{0.211}$            & $4.767_{0.976}$  &  $0.195_{0.035}$           \\
    \bottomrule
    \end{tabular}
\end{table}

\section{Conclusion}
In this paper, we proposed the Conditional Monge Gap, a novel approach for learning OT maps conditionally that was
illustrated on single-cell perturbation response prediction for single or multiple conditions, such as one or multiple drugs, drug dosage, and their combination. 
Our proposed framework can easily be applied on other covariates such as time or genetic perturbations or CAR-T cell therapy, where concurrent work has already reported compelling results in predicting single-cell response for unseen CAR variants with the conditional Monge Gap~\citep{driessen2024modeling}. 

Notably, our lightweight, architecture-agnostic approach extrapolates well to unseen drug treatments.
Our results were especially encouraging when considering effect-driven embeddings (MoA) but we showed that 
the performance gap to a structure-driven embedding (RDkit fingerprint) narrows the more conditions we use.
This large-scale experiment with over 700 conditions (each with around 1000 cells) revealed a promising path to successful OOD predictions and highlight the superiority of conditional OT over autoencoder based methods. 

Future work could incorporate unbalancedness to combat outliers or undesired distribution shifts \citep{lubeck2022neural,eyring2024unbalancedness,klein2024genot} or investigate the impact of larger or more expressive architectures than the MLP utilized in the Monge Gap, e.g., by integrating it with the emerging single-cell foundation models~\citep{cui2024scgpt,yang2022scbert}. 
Additionally, flow matching has now become a popular option for solving optimal transport problems that can incorporate unbalancedness and stochasticity, which allows sampling from the OT map \citep{klein2024genot}. 

\section*{Code availability}
The source code for reproducing the experiments is available at:~\url{https://github.com/AI4SCR/Conditional-Monge} and can be installed from \href{https://pypi.org/project/cmonge/}{PyPI} as~\texttt{cmonge}.

\section*{Acknowledgements}
We thank Juan Gonzalez-Espitia for helpful discussions.
J.B. would like to acknowledge support from the EU project Fragment-Screen (grant agreement ID 101094131). This project received funding from the European Union’s Horizon 2020 research and innovation program under the Marie Skłodowska-Curie grant agreement no. 955321.
\newpage

\section{Methods}
\subsection{Background}
\label{sec:ot_background}
The Monge formulation of OT seeks an optimal map $T:\mathbb{R}^d\to\mathbb{R}^d$ between probability measures $(\mu, \nu)\in \mathcal{P}(\mathbb{R}^d)\times\mathcal{P}(\mathbb{R}^d)$, s.t. $T$ pushes forward $\mu$ onto $\nu$, while minimizing a displacement cost:
\begin{align}
     T^*:=\arg \inf_{T\sharp\mu =\nu}\int_{\mathbb{R}^d} c(x,T(x)) dx.
\end{align}
In practice, measures are often data samples $\mu=\frac{1}{n}\sum_{i=1}^n\delta_{x_i}$ and $\nu= \frac{1}{n}\sum_{j=1}^n\delta_{y_j}$, where $\delta$ is the Dirac delta function. In that case, the OT problem is solved through the entropic regularized Kantorovich relaxation, which reads:
\begin{gather}\label{eq:kantor}
    W_{\epsilon}(\mu, \nu):=\min_{P\in U_n} \langle P, C \rangle + \epsilon H(P), 
\end{gather}
\begin{gather}
    U_n=\{P\in \mathbb{R}^{n\times n}_+ : P1_n=\frac{1}{n}1_n, P^T1_n=\frac{1}{n}1_n\}
\end{gather}
where $H(P)=-\sum_{i,j}P_{ij}\log(P_{ij})$ is the entropy of coupling matrix $P$ and with $\epsilon>0$. $P$ describes the amount of mass flowing between the samples. $U_n$ is the set of all possible couplings that satisfy the marginals. $C$ is the cost matrix $C_{i,j}=c(x_i, y_j)$ and represents the cost of moving $i$ to $j$ and: 
\begin{align}
 \langle P, C \rangle = \sum_{i,j} P_{i,j}C_{i,j}
\end{align}
We can construct a differentiable loss function, the Sinkhorn divergence, by debiasing the objective \cite{genevay2018learning}, such that $W_{\epsilon}(\mu, \mu)=0$ holds with the modification
\begin{align}
    \Delta_{\epsilon}(\mu, \nu)= W_{\epsilon}(\mu, \nu) - \frac{1}{2}(W_{\epsilon}(\mu, \mu)+W_{\epsilon}(\nu, \nu)). \label{eq:sink_div}
\end{align}

\subsection{Evaluation settings}
\label{subsec:eval_settings}
We evaluated the Monge Gap and the Conditional Monge (\texttt{CMonge}) in different settings. The basic Monge models are unaware of conditional contexts and thus are a single model per condition. In cases where a Monge model is trained on multiple conditions, it does not receive any conditional information and therefore treats all conditions as equal. 

All in-distribution (ID) models are trained on data of all conditions, using an 80/20 train/test split for each condition. The evaluation conditions were already seen during training. For all Monge models, this means that the dose and drug cannot be distinguished. Conversely, for all CMonge models, the dose, and, when applicable, the drug, is encoded and this information is given to the models. 

In the out-of-distribution (OOD) setting, the Monge and CMonge models are evaluated on held-out conditions and trained on all other conditions. Unless specified, we employed a leave-one-out setting where we trained $n$ models for $n$ held-out conditions, always leaving one condition out. As in the ID setting, the Monge models do not distinguish conditions in the training or evaluation setting, whereas CMonge can be conditioned on dosage alone or drug and dosage. 

First, we tested the model's ability to condition on the dosage and to generalize to unseen dosages. Therefore, the CMonge models were conditioned on dosage (Dose) in both an ID and OOD setting. Then, we tested the model's ability to condition on drugs and generalize to unseen drugs by holding out all dosages of one drug during training. The \texttt{CMonge} models were conditioned on both drug and dosage (DrugDose) in an ID and OOD setting. Lastly, we investigated model performance on conditioning on DrugDosage with more training drugs, both in an ID and OOD setting. In the OOD setting, we increased the OOD conditions by leaving 9 drugs out instead of one. 

We evaluated with the $R^2$ metric between the perturbed and predicted feature means, the Wasserstein distance (\autoref{eq:kantor}), and the Maximum Mean Discrepancy (MMD). The metrics are calculated on batches of observations sampled from the test set and we report the mean across these batches. The experiments used the same hyperparameters, except for the number of optimization steps and the latent encoding of the features and context (for more details, see Methods \ref{sec:app_model_size} \& \ref{cond_appendix}).
For the SciPlex dataset, we trained and evaluated the Conditional Monge Gap in the following scenarios:
\setlist{nolistsep}
\begin{enumerate}[leftmargin=*,noitemsep]
    \item \texttt{Monge}: As a hypothetical upper bound on performance, we fit separate Monge Gap models, one per drug-dosage pair. These models do not have any context (cf.~\citet{uscidda2023monge}).
    \item \texttt{Monge-Drug} / \texttt{Monge-DrugDose}: To motivate the contextual settings, we fit a Monge Gap model that is trained on multiple conditions but unaware of conditional information and evaluate it on conditions seen during training. 
    \begin{itemize}
        \item Monge-Dose-ID: a homogeneous model for each drug, using data from all four dosages. 
        \item Monge-DrugDose-ID: one model on all conditions (all drug-dosage pairs). 
        \item Monge-Dose-OOD: a model for each drug and left out different dosages during training. 
        \item Monge-DrugDose-OOD: a model trained on all conditions but the dosages of the held-out drug(s). 
    \end{itemize} 
    \item \texttt{CMonge-Dose-ID} / \texttt{CMonge-Dose-OOD}: We fit conditional models for each drug with the scalar dose as context. The ID setting sees all dosages during training. For the OOD setting, we left out different dosages during training, thus creating interpolation and extrapolation settings. 
    \item \texttt{CMonge-DrugDose-RDKit / CMonge-DrugDose-MoA}: A single model fitted to all data, conditioned on drug and dosage context. To encode the drug, we compare fingerprints (\texttt{RDKit}) to a data-driven approach (\texttt{MoA}).
    \begin{itemize}
        \item CMonge-DrugDose-x-ID: All conditions are seen during training
        \item CMonge-DrugDose-x-OOD: All dosages of one or more drugs are held during training for evaluation. 
    \end{itemize}      
\end{enumerate}

\subsection{Model sizes}
\label{sec:app_model_size}
To allow CMonge to learn from this increased amount of data and to make comparison to chemCPA fair, we increased the model size of CMonge and the embedding size of the gene expression and the context information. The drug and data embedding was increased from 50 to 100 dimensions, increasing the gene-expression autoencoder from 1.60M to  1.65M parameters. The four fully connected layers of CMonge were increased from four times 64 to twice 256 and twice 512, resulting in a parameter increase from 23K to 560/580K (for MoA/RDkit embedding, respectively). For comparison, chemCPA, which has one model for embedding the data, context information, and learning the perturbation effect, has around 1.37M parameters. 

\subsection{Hyperparameters of the conditional experiments}
\label{cond_appendix}

In all experiments, we use the AdamW optimizer \citep{loshchilov2017decoupled}, with initial learning rate $10^{-4}$ and weight decay regularization ${10^{-5}}$. Unless stated otherwise, both the encoder and decoder consist of two hidden layers of $512$ dimensions each. The $50$-dimensional latent representation is learned through $50$ epochs with a batch size of $256$. The Monge network is built out of 4 hidden layers with $64$ neurons each. The dose and drug embedders are parameterized with one dense layer. The Euclidean distance is used as displacement cost and the Monge Gap regularizer is set to $\lambda=10^{-2}$. During the OT training phase, we repeatedly sample a batch of $256$ observations from the source and target distributions for $1000$ iterations for local models (without condition, or conditioned on dosage), while $10000$ iterations for the global models (RDKit and MoA). And $500000$ iterations for the bigger models, trained all drugs in the SciPlex dataset. Each batch only contains samples from one context, which is uniformly sampled. All models are implemented using the \textsc{\texttt{OTT-JAX}} package \citep{cuturi2022optimal}. 

\paragraph{Conditional 4i dataset}
\label{sec:cond_4i_treatments}
For the conditional Monge 4i experiments, combinatorial therapies (conditions with two or three drugs) were handled similarly for the RDkit and the MoA embeddings. For each drug in the condition, the embedding was computed for the MoA based on the single drug condition if possible. Then, the embeddings from all drugs in the condition were passed through the same dense layer, with the same parameters for each single drug embedding. For the MoA embedding, some single-drug embeddings were missing, as the effect of the single drug was not measured. These drugs were then omitted from calculating the condition embedding. Additionally, we left out the conditions `vemurafenib-cobimetinib` since neither drug was measured in isolation and therefore no MoA embedding was possible for either single drug and the condition `pomalidomide-carfilzomib-dexamethasone` as only dexamethasone was present as single treatment.

\subsection{Benchmarking}
We first benchmarked different state-of-the-art methods for unconditional perturbation modeling.
\paragraph{4i dataset}
For the non-conditional experiments, we trained each method on each of the 35 therapies with an 80/20 train/validation split. We note that the original scGen~ \citep{lotfollahi2019scgen} relies on a Variational formulation ~\citep{kingma2013auto}. Instead, in our experiments, we follow the setup in~ \citet{bunne2022supervised}, i.e., we utilized a vanilla AE, and both the encoder and the decoder are parameterized with fully connected layers.
The results in ~\autoref{tab:4i_means_std} and ~\autoref{fig:4i_bar} confirm the finding by ~\cite{uscidda2023monge}, i.e., the Monge Gap achieves the overall best result with respect to each of the evaluation metrics and also shows lower standard deviation. In ~\autoref{fig:4i_bar}, the subplot on the Wasserstein distance clearly shows that the Monge model (which directly optimizes this metric) performs consistently, regardless of the perturbation, while the autoencoder is struggling to capture perturbation effects in some cases. On the other hand, the ICNN results are only skewed by one outlier. Remarkably, the optimal transport-based models are outperforming the autoencoder on MMD, $R^2$, and Drug Signatures, even though they are trained to optimize the primal and dual OT loss.

\paragraph{SciPlex}
For each of the nine drugs and each of the four dosages, we fitted a different model, resulting in 36 models per method. We included the identity mapping as a baseline, which simply predicts the unperturbed cell states. ~\autoref{table:benchmark} and~\autoref{fig:line-benchmark} show the performance of the different methods. Our main evaluation metric is still the Coefficient of Determination of the HVG feature means $(R^2)$, but we also report the entropy-regularized Wasserstein distance, which is closely related to the objective function of the ICNN and Monge models. The ICNN and Monge models used the $50$ dimensional latent representation learned by the autoencoder. 

All model predictions were decoded and evaluated in the cell space on the $50$ HVGs. We observe that the neural optimal transport-based solvers significantly outperform the autoencoder-based approach. 

The results of~\autoref{fig:scatter-benchmark} show that, although the ICNN-based solver slightly outperforms the Monge-based counterpart, their results are highly correlated, and they have almost identical performance for the Wasserstein distance. Combining this with the fact that we are expanding upon the Monge-based methodology, we did not include ICNN-based benchmarks in the main work.

\bibliography{references}
\bibliographystyle{bibstyle}
\newpage
\appendix
\section{Appendix}
\renewcommand{\thefigure}{A\arabic{figure}}
\renewcommand{\thetable}{A\arabic{table}}
\setcounter{figure}{0} 
\setcounter{table}{0} 
\subsection{Supplemental figures}

\begin{figure}
    \centering
    \includegraphics[width=1\textwidth]{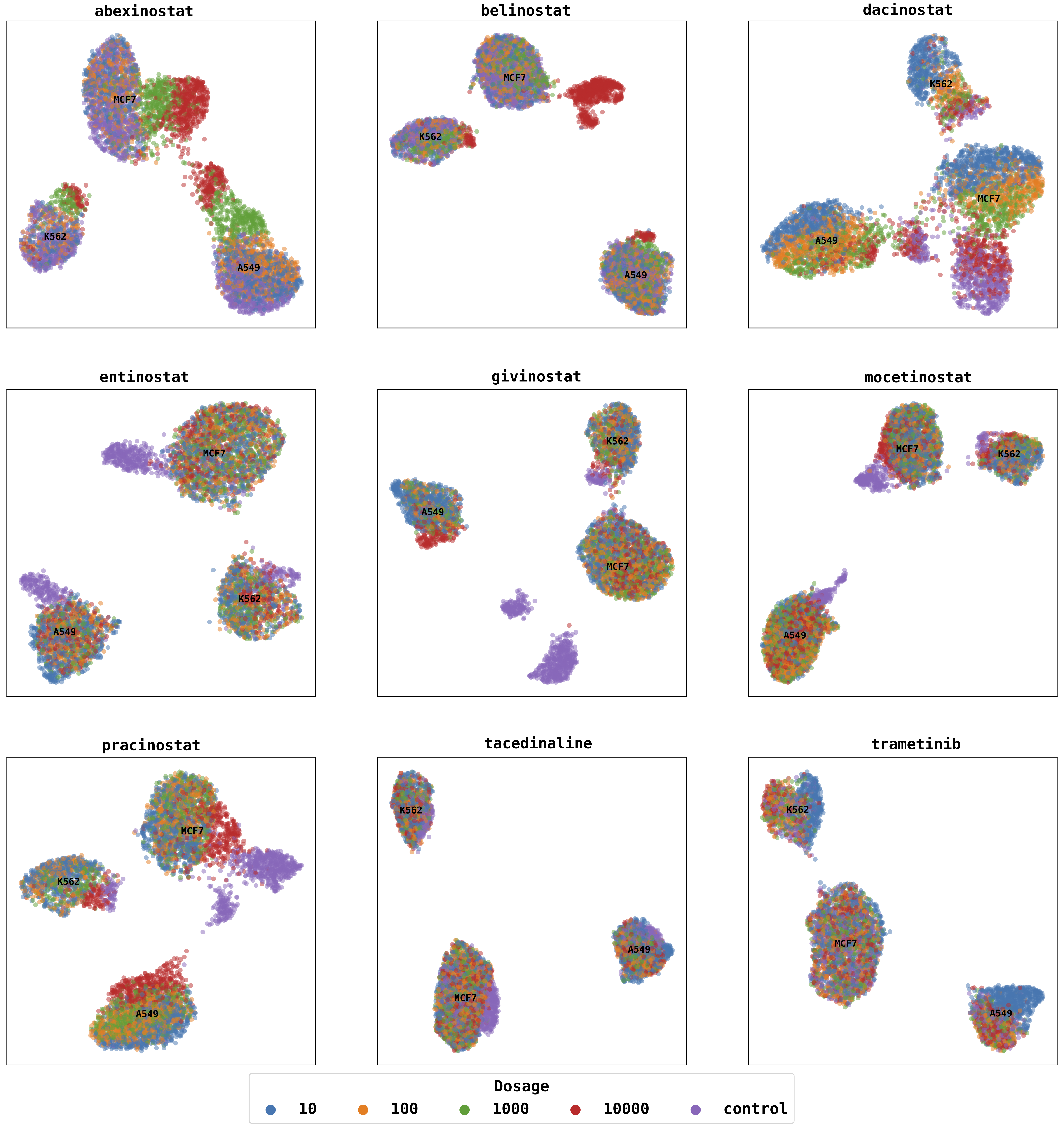}
    \caption{UMAP projection of the $1000$-dimensional feature space, filtered on control cells and cells treated with different dosages. We can observe a greater perturbation effect with higher dosage. Moreover, the three clusters are associated with the three different cell types (see black text).}
    \label{fig:umap}
\end{figure}

\begin{figure}
    \centering
    \includegraphics[width=1\textwidth]{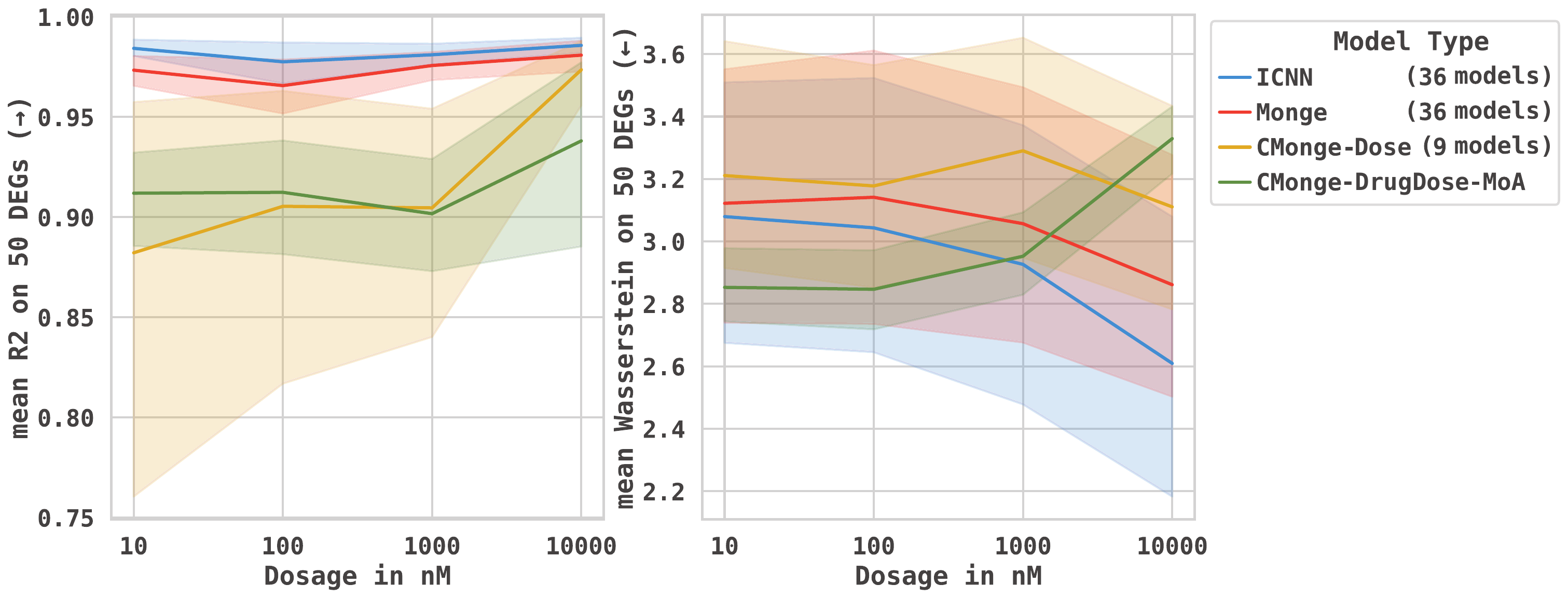}
    \caption{Comparison between the unconditional OT-based models and the conditional counterparts on the ScipPlex dataset in the in-distribution setting based on the $R^2$ of feature means (left) and the Wasserstein distance (right).}
    \label{fig:line36}
\end{figure}

\begin{figure} [!htp]
    \centering
    \includegraphics[width=1\textwidth]{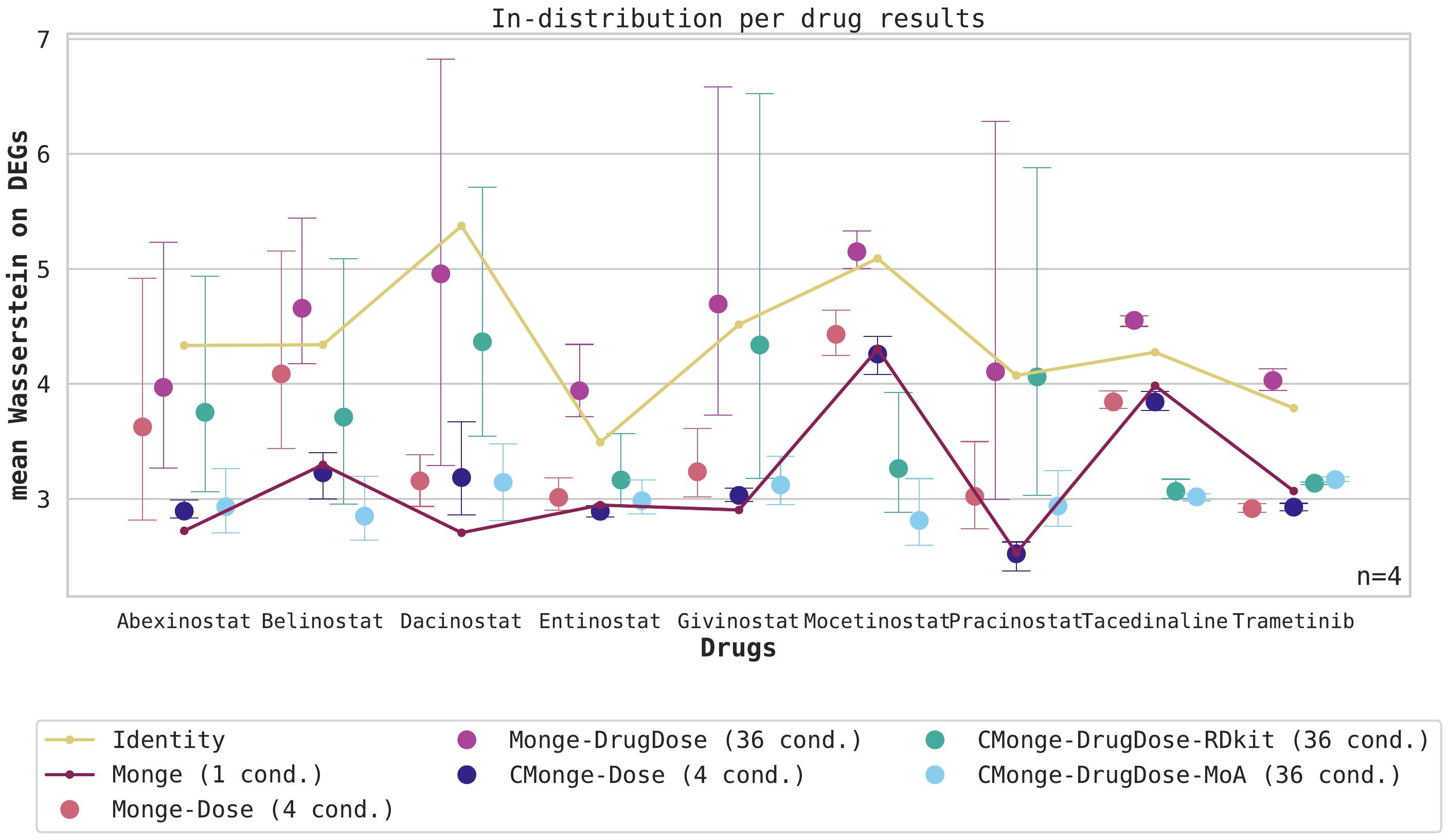}
    \caption{Comparison of the different conditional and unconditional Monge methods for the ID setting. Results are grouped by drug and show the \textbf{Wasserstein distance}. Each point represents the mean performance of the model over the four dosages. Error bars represent the 95\% confidence interval. Note that CMonge-Dose and Monge-Dose are models per-drug (trained on only one drug), whereas CMonge-DrugDose are pan-condition models where one model is trained on all drugs and dosages. Monge is one model per condition (36 in total). CM: CMonge}
    \label{fig:point_w}
\end{figure} 

\begin{figure}[!htb]
    \centering
    \includegraphics[width=1\textwidth]{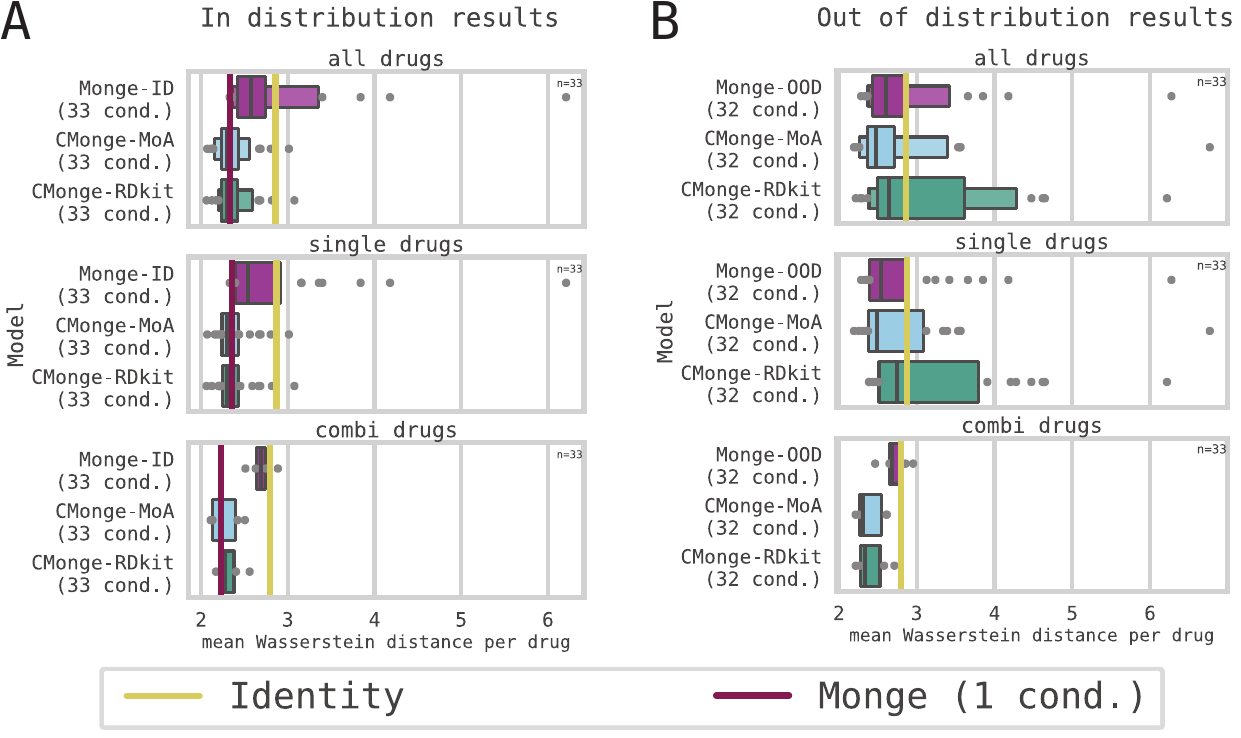}
    \caption{
        Comparison of different Monge, Conditional Monge (CMonge), and the Identity models using the mean MMD. Panels show overall performance, performance for single treatments, and combinatorial treatments.
        A) In distribution results. The Monge model is trained as one model that sees all conditions, but is not aware of conditional information. The CMonge models are likewise trained on all conditions, but do get conditional information with the Mode-of-Action (MoA) or RDkit embedding.  
        B) Out of distribution (OOD) results. All models are trained in a leave-one-drug-out setting. This means that each boxenplot shows the result over 33 drugs, from 33 models. The Monge model is also trained in a leave-one-drug-out setting, but does not incorporate conditional information.
        }
    \label{fig:4i_EMD}
\end{figure}

\begin{figure}[!htb]
    \centering
    \includegraphics[width=1\textwidth]{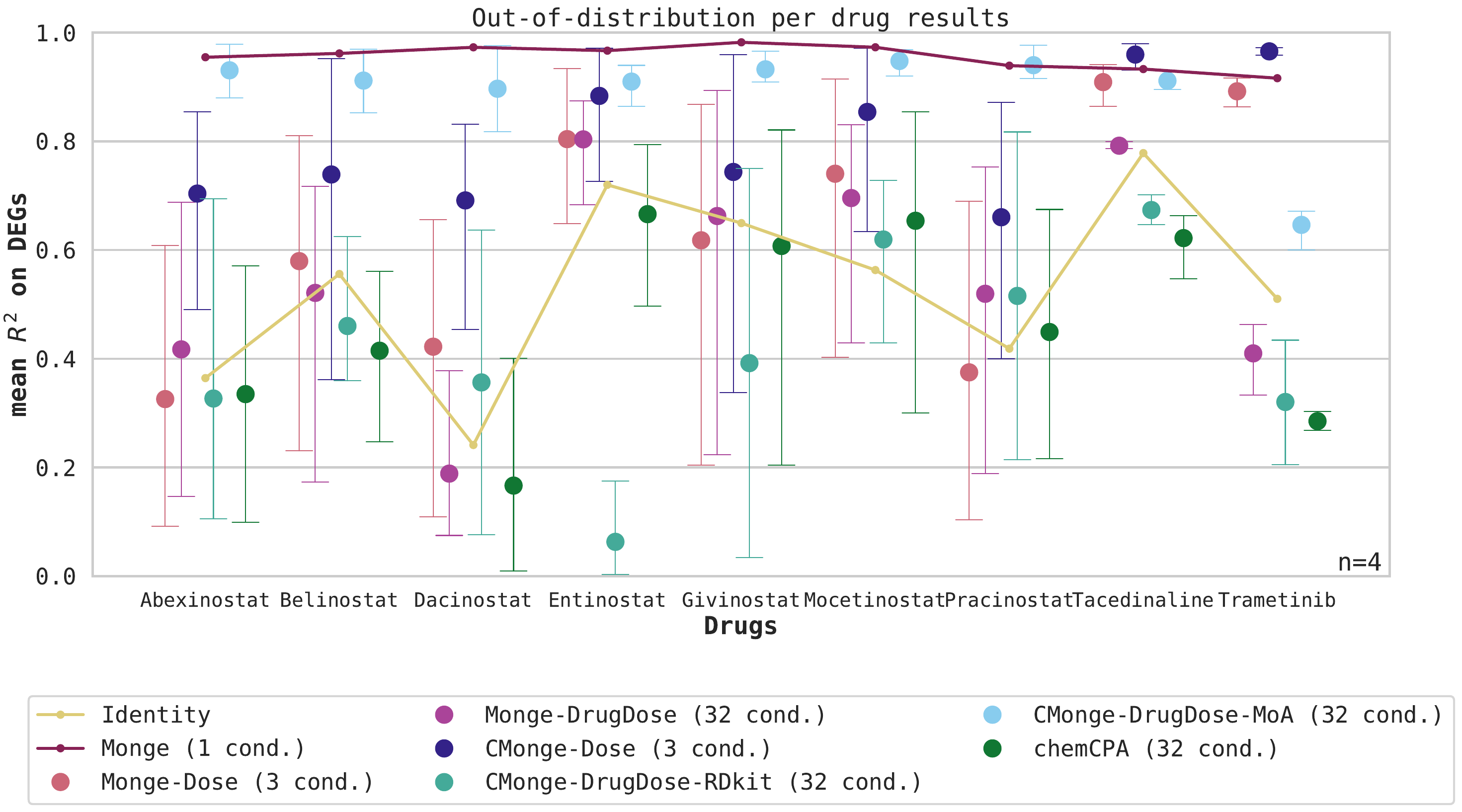}
    \caption{
    Results of the OOD experiments, comparing the Monge Gap, the conditional Monge Gap, chemCPA and the Identity and Monge per condition baselines. Results grouped by drug, using the mean $R^2$ metric. Each point represents the mean performance of the model out of the four dosages, along with the 95\% confidence interval around the mean. Since chemCPA is conditioned on cell line, drug and dosage there are 12 points per estimate. For all other models this is four points.}
\label{fig:ood_sciplex_pointplot_R2}
\end{figure}

\begin{figure} [!htp]
    \centering
    \includegraphics[width=1\textwidth]{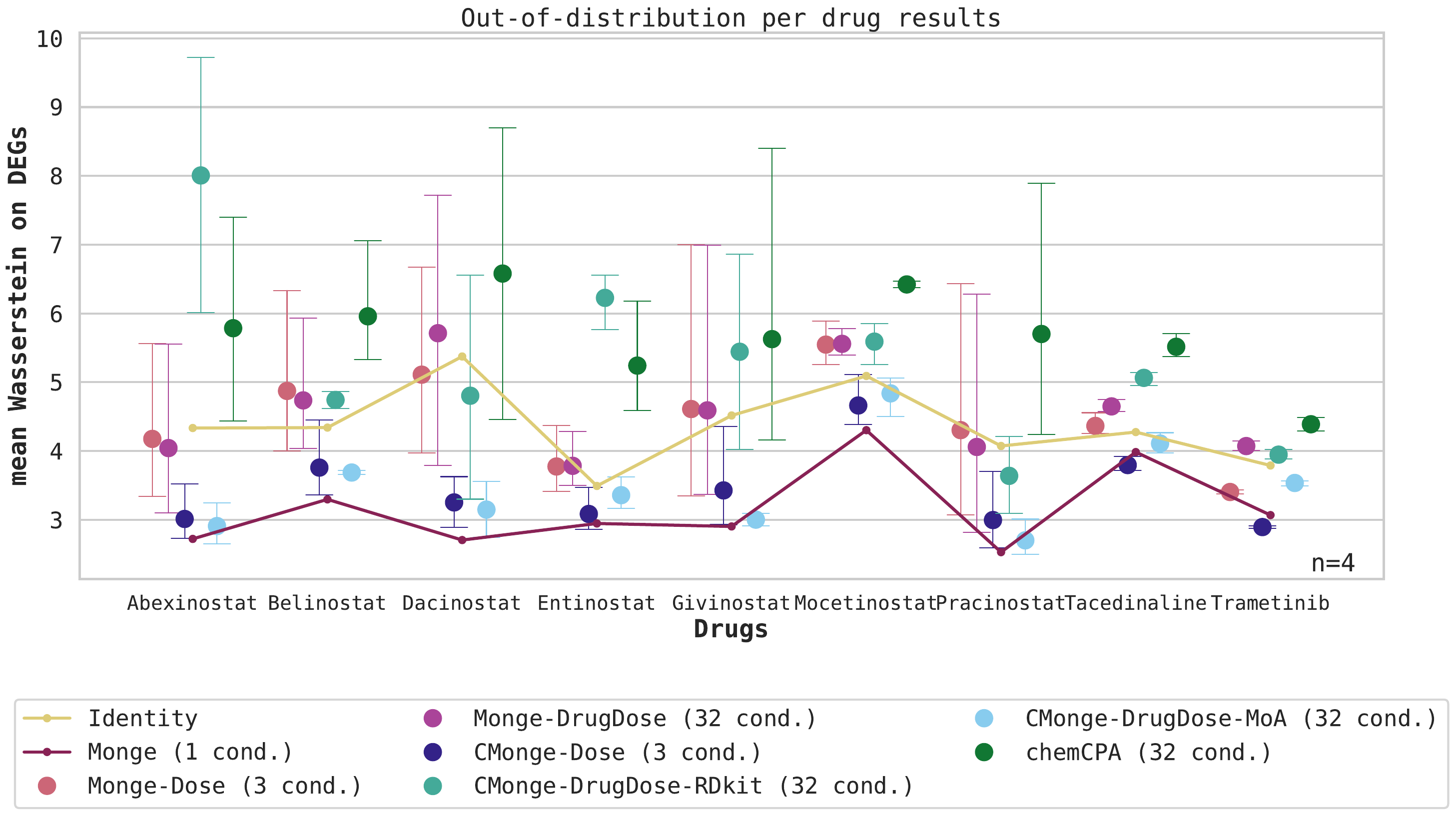}
    \caption{Results of the OOD experiments, comparing the Monge Gap, the conditional Monge Gap, and the Identity and Monge per condition baselines. Results grouped by drug, using the mean Wasserstein distance. Each point represents the mean performance of the model out of the four dosages, along with the 95\% confidence interval around the mean. chemCPA results are not shown here, as chemCPA predicts a mean and variance per gene and per cell, so not directly the gene expression. Therefore, the Wasserstein distance cannot be computed in a straightforward manner.}
    \label{fig:point_w_ood}
\end{figure} 

\begin{figure}[!htb]
    \centering
    \includegraphics[width=1\textwidth]{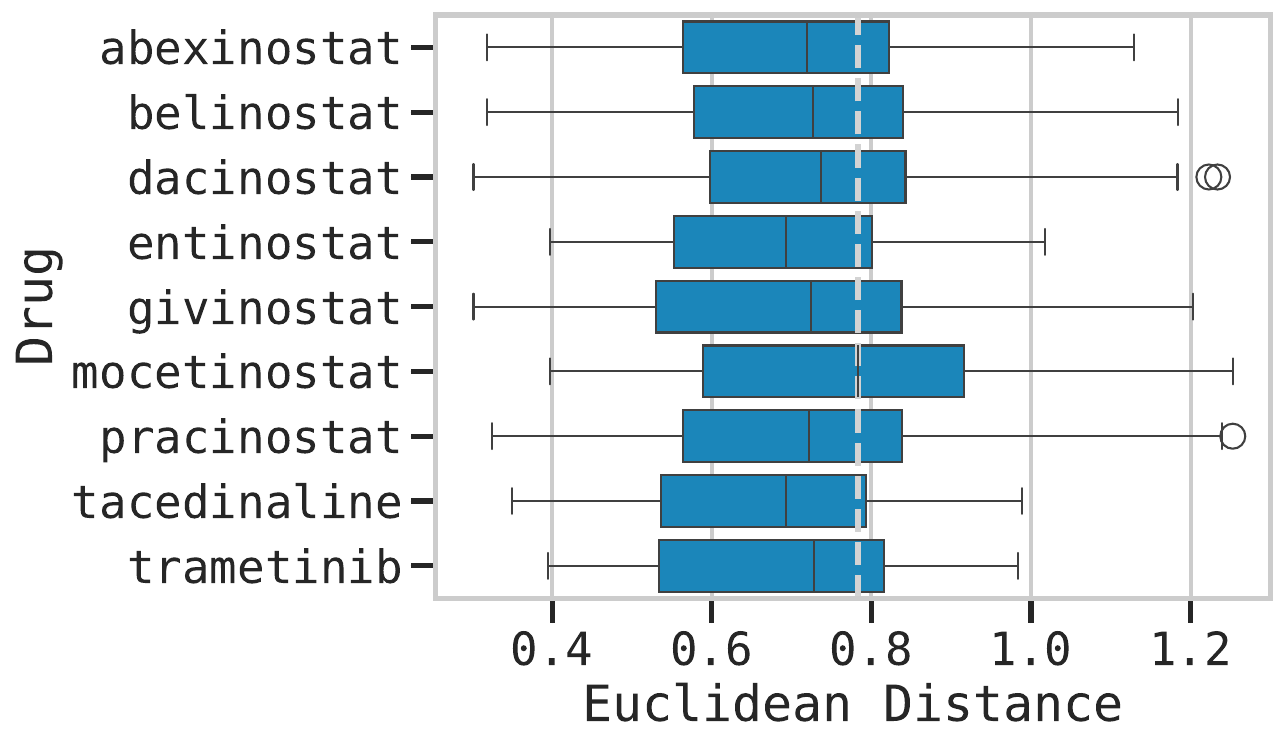}
    \caption{Euclidean distance in Mode of Action (MoA) embedding space. Since we held out all dosages of one drug in the CMonge-DrugDose-OOD-MoA experiments, the distances between dosages within the same drug are not included. Mocetinostat has the highest distance to the other conditions. For pairwise distances see \autoref{fig:MoA_embed_dist}. The dashed line shows the median distance of all mocetinostat conditions to all non-mocetinostat conditions.}
    \label{fig:MoA_embed_dist_box}
\end{figure}

\begin{figure}[!htb]
    \centering
    \includegraphics[width=1\textwidth]{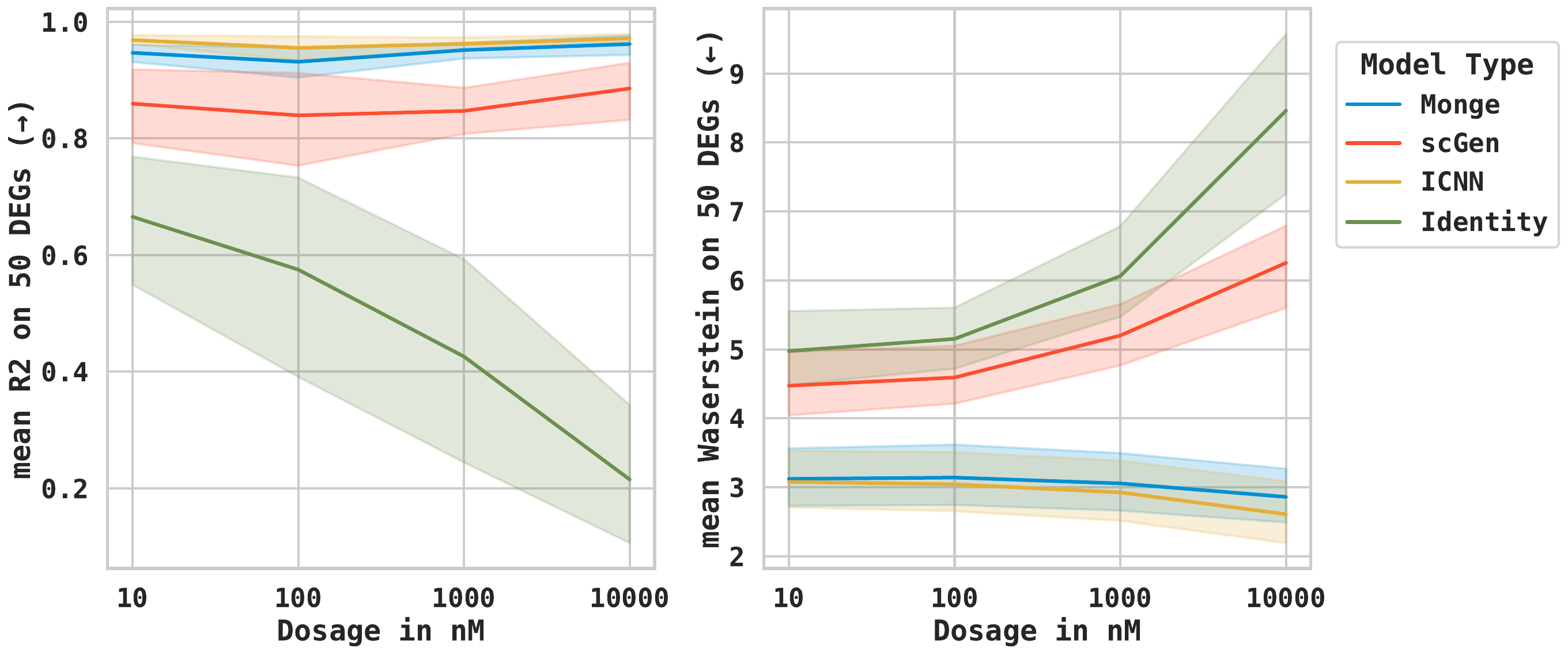}
    \caption{Performance of the benchmarked models for the 4 different dosages, averaged over 9 drugs.}
    \label{fig:line-benchmark}
\end{figure}

\begin{figure}
    \centering
    \includegraphics[width=120mm]{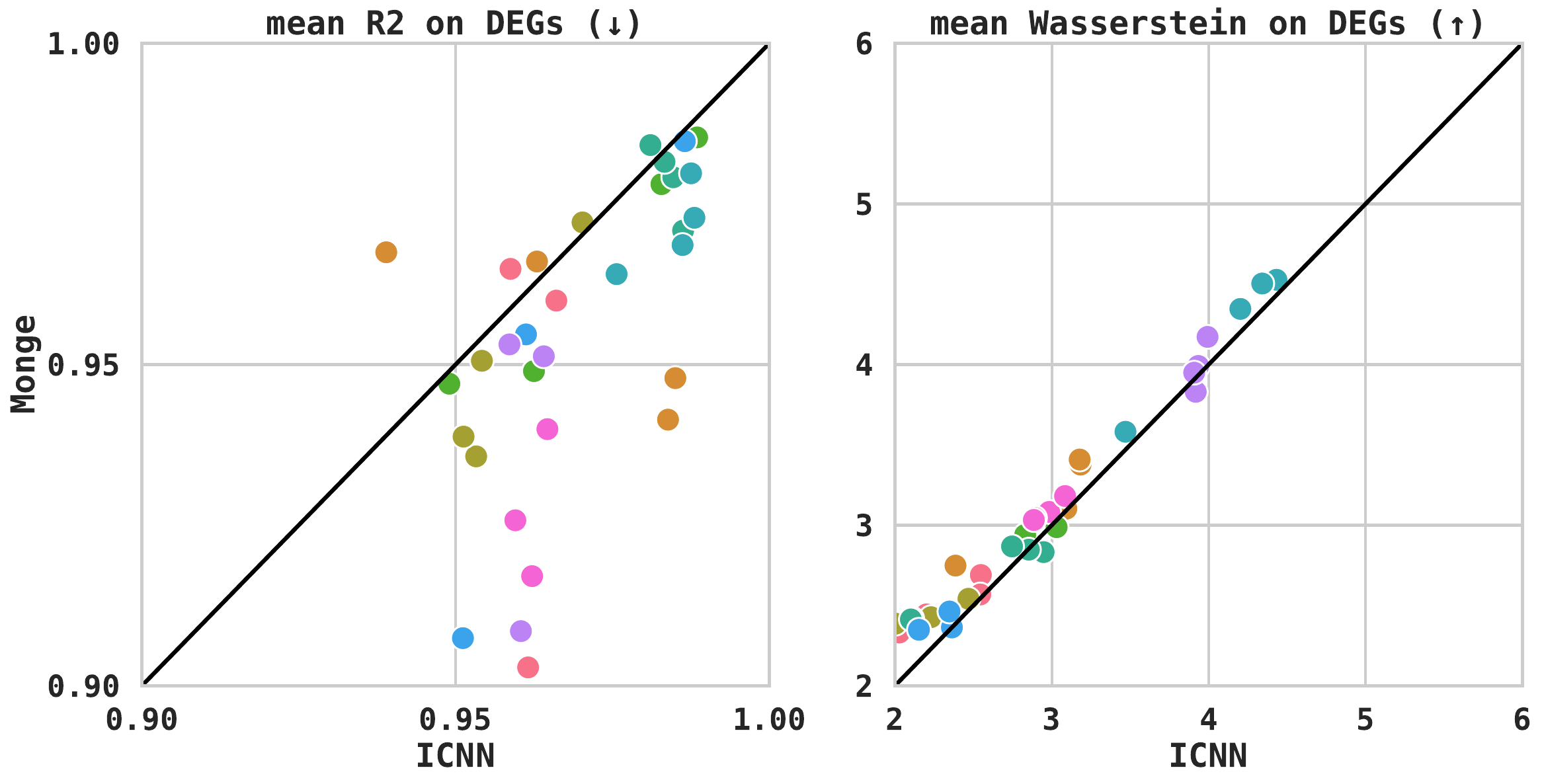}
    \caption{Comparison of neural optimal transport solvers, the scatter plot consists of $(x_i, y_i)$ points, where $x_i$ represents the target metric ($R^2$ or Wasserstein) obtained by the ICNN solver, and $y_i$ is the performance of the corresponding Monge model on the same drug-dose split. Each drug is denoted with a different color. In the case of the $R^2$ metric, each time an $(x_i, y_i)$ point is under the $x=y$ line, the ICNN outperforms the Monge model. }
    \label{fig:scatter-benchmark}
\end{figure}

\begin{figure}[!htb]
    \centering
    \includegraphics[width=1\textwidth]{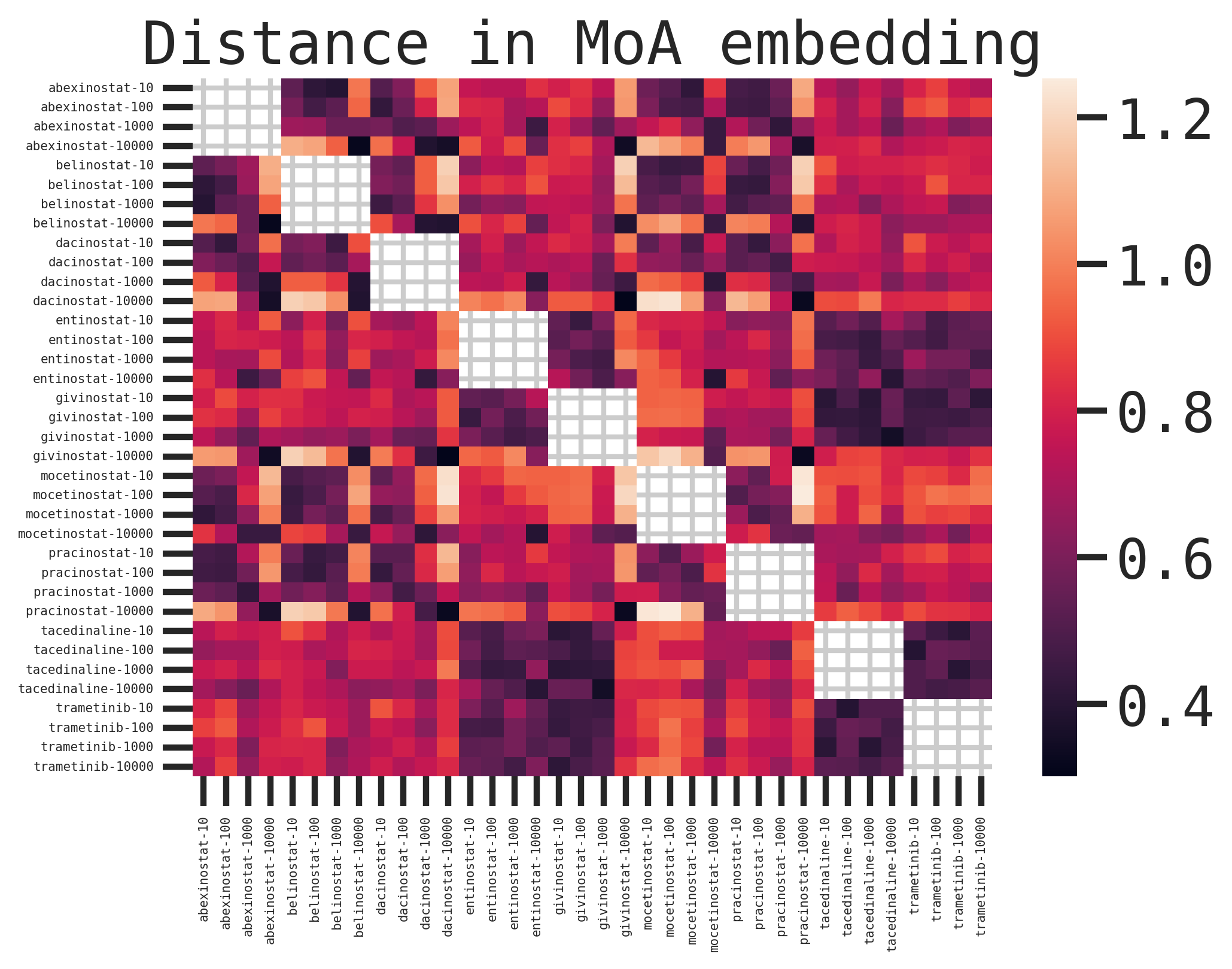}
    \caption{Euclidean distance in Mode of Action (MoA) embedding space. Since we held out all dosages of one drug in the CMonge-DrugDose-ood-MoA experiments, the distances between dosages within the same drug are not calculated. Mocetinostat has a high distance to many other drugs and dosages, especially for the three lower dosages.}
    \label{fig:MoA_embed_dist}
\end{figure}

\begin{figure}
    \centering
    \includegraphics[width=120mm]{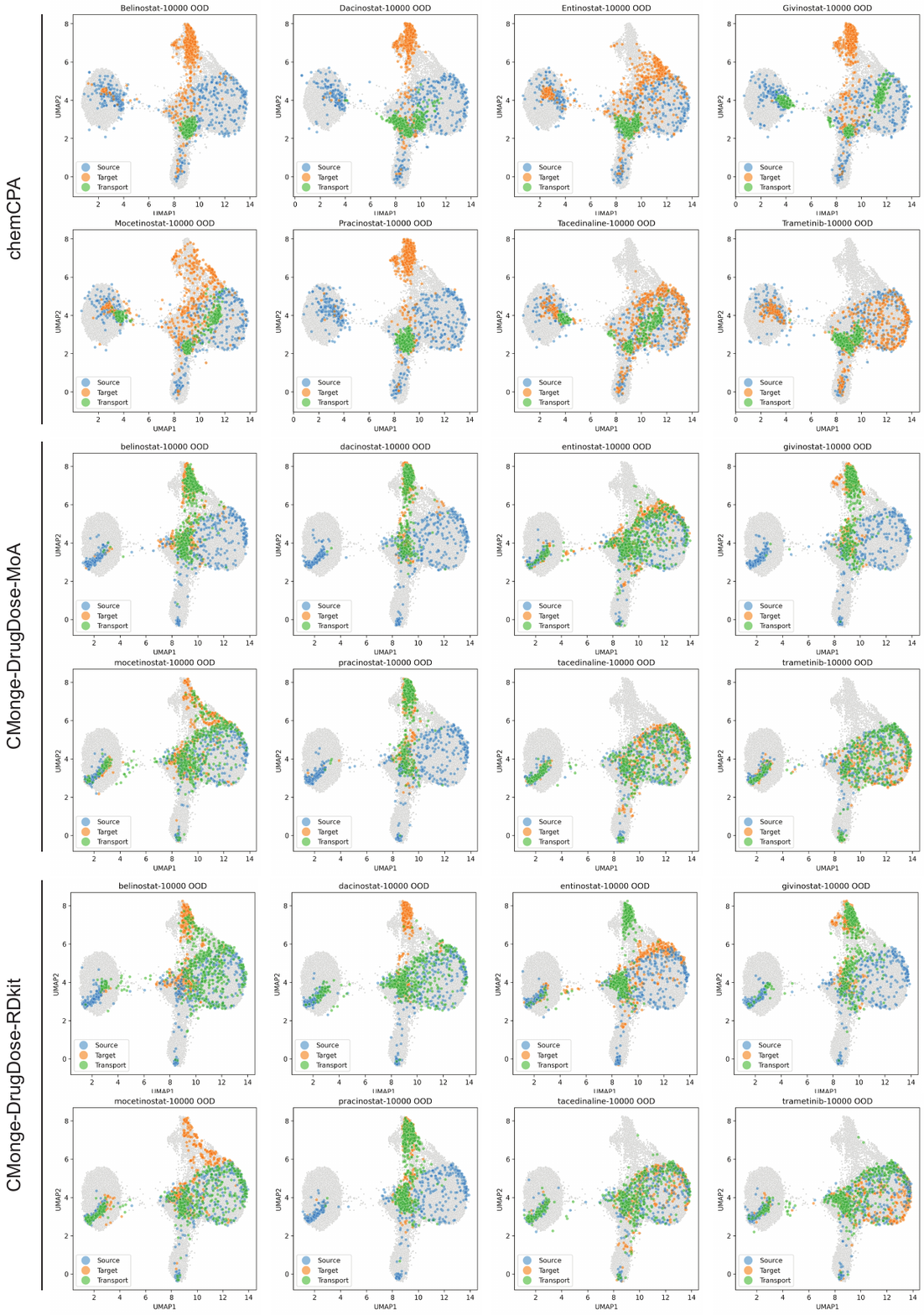}
    \caption{Drug-wise comparison between chemCPA~\citep{hetzel2022predicting} and the two global conditional models, where we condition based on drug and dosage as well.}
    \label{fig:extra_umaps}
\end{figure}


\FloatBarrier
\subsection{Supplemental tables}

\begin{table}[!htb]
  \centering
  \footnotesize
  \caption{Evaluation of perturbation prediction on the 4i dataset. Average performance is reported over the 35 treatments, along with the standard deviation in the 48-dimensional feature space.}
  \label{tab:4i_means_std}
  \begin{tabular}{lccccc}
    \toprule
    Model & Wasserstein & MMD & $R^2$ & Drug Signature & Sinkhorn Div \\
    \midrule
    {\mycirc[blue]}Monge & \bftab 2.345 $\pm$ 0.219 & \bftab 0.009 $\pm$ 0.001 & \bftab 0.901 $\pm$ 0.087 & \bftab 0.333 $\pm$ 0.067 & \bftab1.812 $\pm$ 0.221 \\
    {\mycirc[red]}AE & $2.455 \pm 0.396$ & $0.024 \pm 0.014$ & $0.804 \pm 0.110$ & $0.523 \pm 0.150$ & $1.921 \pm 0.398$ \\
    {\mycirc[violet]}ICNN & $2.632 \pm 0.512$ & $0.009 \pm 0.007$ & $0.878 \pm 0.104$ & $0.433 \pm 0.406$ & $2.097 \pm 0.515$ \\
    {\mycirc[green]}Identity & $2.881 \pm 0.736$ & $0.033 \pm 0.027$ & $0.309 \pm 0.209$ & $1.285 \pm 0.891$ & $2.347 \pm 0.738$ \\
    
    \bottomrule
  \end{tabular}
\end{table}

\tabcolsep=0.10cm
\begin{table}[!htb]
\centering\footnotesize
\caption{Evaluation of conditional and unconditional drug and dose experiments. Results are compared based on the Coefficient of Determination between the predicted and target feature means ($R^2$). Results show average and standard deviation across 36 conditions (9 drugs, each with 4 dosages).}
\label{table:drug_dose_id}
\begin{tabular}{lccccccc}
    \toprule
    Model           & conditions& \multicolumn{2}{ c }{Context}  &   \multicolumn{4}{ c }{Dosage (nM)}\\
                    & seen      & Drug & Dose                    &10 & 100 & 1000 & 10000    \\
    \midrule 
    \midrule
    Identity                    &      &             &           & $0.747_{0.126}$           & $0.654_{0.260}$           & $0.503_{0.332}$           & $0.227_{0.212}$           \\
    Monge                       &  1   &             &           & $\mathbf{0.950_{0.020}}$  & $\mathbf{0.935_{0.042}}$  & $\mathbf{0.960_{0.025}}$  & $\mathbf{0.978_{0.029}}$  \\
    Monge-DrugDose-ID           &  36  &             &           & $0.699_{0.181}$           & $0.712_{0.131}$           & $0.659_{0.255}$           & $0.292_{0.318}$           \\
    CMonge-DrugDose-RDkit-ID    &  36  & \checkmark & \checkmark & $0.619_{0.268}$           & $0.690_{0.172}$           & $0.868_{0.056}$           & $0.352_{0.320}$           \\
    CMonge-DrugDose-MoA-ID      &  36  & \checkmark & \checkmark & $0.912_{0.039}$           & $0.912_{0.046}$           & $0.902_{0.048}$            & $0.938_{0.079}$  \\
    \bottomrule
    \end{tabular}
\end{table}

\begin{table}[!htb]
\centering\footnotesize
\caption{Evaluation of drug effect perturbations, treated with 9 different drugs. Results are compared based on the \textbf{Wasserstein distance} between the predicted and target samples. The average and standard deviation are reported for the 9 experiments per model.}
\label{table:main_w}
\begin{tabular}{lcllcccc}

    \toprule
    Model                   & conditions     &\multicolumn{2}{ c }{Context} &\multicolumn{4}{ c }{Dosage (nM)}\\
                            &    &   Drug    &   Dose    &   10              &   100             &   1000            &   10000   \\
    \midrule
    \midrule 
    Identity                &    &           &           & $3.444_{0.811}$   & $3.511_{0.683}$   & $4.103_{1.000}$   & $6.400_{1.720}$\\
    Monge                   &    &           &           & $3.120_{0.693}$   & $3.162_{0.709}$   & $3.164_{0.603}$   & $3.200_{0.465}$ \\
    \midrule
    \midrule
    Monge-Dose-ID           & 4  &           &           & $3.326_{0.532}$   & $3.291_{0.512}$   & $3.175_{0.516}$   & $4.129_{0.981}$ \\
    Monge-DrugDose-ID       & 36 &           &           & $3.941_{0.764}$   & $3.876_{0.633}$   & $4.090_{0.727}$   & $5.894_{1.442}$ \\
    \midrule
    \midrule
    Monge-Dose-OOD          & 3  &          &           & $4.075_{0.761}$   & $3.869_{0.681}$   & $3.775_{0.725}$   & $6.130_{1.588}$ \\
    Monge-DrugDose-OOD      & 32 &          &           & $3.880_{0.913}$   & $6.260_{1.682}$   & $4.230_{1.045}$   & $3.940_{0.850}$ \\
    \midrule
    \midrule
    CMonge-Dose-ID          & 4  &           &\checkmark & $3.211_{0.568}$  & $3.178_{0.586}$   & $3.290_{0.564}$   & $3.111_{0.544}$ \\
    CM-DrugDose-RDKit-ID    & 36 &\checkmark &\checkmark & $3.229_{0.291}$  & $3.144_{0.182}$   & $3.058_{0.239}$   & $5.171_{1.640}$ \\
    CM-DrugDose-MoA-ID      & 36 &\checkmark &\checkmark & $2.853_{0.193}$	& $2.847_{0.209}$   & $2.953_{0.213}$   & $3.329_{0.182}$ \\
    \midrule
    \midrule
    CMonge-Dose-OOD         & 3  &           &\checkmark & $3.255_{0.592}$  & $3.149_{0.600}$   & $3.197_{0.543}$   & $4.122_{0.745}$ \\
    CM-DrugDose-RDKit-OOD   & 32 &\checkmark &\checkmark & $4.853_{1.443}$  & $5.540_{2.046}$   & $5.793_{1.667}$   & $4.907_{1.059}$ \\
    CM-DrugDose-MoA-OOD     & 32 &\checkmark &\checkmark & $3.368_{0.813}$  & $3.359_{0.819}$   & $3.516_{0.676}$   & $3.659_{0.439}$ \\
    \bottomrule
    \end{tabular}
    \end{table}

\tabcolsep=0.10cm
\begin{table}[!htb]
\centering\footnotesize
\caption{Evaluation of conditional drug and dose experiments. Results are compared based on the Coefficient of Determination between the predicted and target feature means ($R^2$). Results show average and standard deviation across all drugs.}
\label{table:large_drug_dose_id}
\begin{tabular}{lccccccc}
    \toprule
    Model           & conditions& \multicolumn{2}{ c }{Context}  &   \multicolumn{4}{ c }{Dosage (nM)}\\
                    & seen      & Drug & Dose                    &10 & 100 & 1000 & 10000    \\
    \midrule 
    \midrule
    CMonge-DrugDose-MoA-ID      &  748 & \checkmark & \checkmark & $\mathbf{0.917_{0.034}}$           & $0.919_{0.029}$           & $0.909_{0.040}$           & $0.890_{0.050}$           \\
    CMonge-DrugDose-RDkit-ID    &  748 & \checkmark & \checkmark & $0.913_{0.037}$           & $\mathbf{0.920_{0.031}}$           & $\mathbf{0.915_{0.033}}$           & $\mathbf{0.901_{0.045}}$           \\
    chemCPA-DrugDoseCellLine-ID &  748 & \checkmark & \checkmark & $0.896_{0.137}$           & $0.849_{0.185}$           & $0.735_{0.231}$           & $0.736_{0.154}$           \\
    \bottomrule
    \end{tabular}
\end{table}

\tabcolsep=0.10cm
\begin{table}[!htb]
\centering\footnotesize
\caption{Evaluation of conditional and unconditional drug and dose out-of-distribution experiments. Results are compared based on the Coefficient of Determination between the predicted and target feature means ($R^2$). The average and standard deviation are reported for the 9 experiments per model. Results averaged across nine drugs. DD: DrugDose}
\label{table:drug_dose_ood}
\begin{tabular}{lccccccc}
    \toprule
    Model (DrugDose)          &  \#       & \multicolumn{2}{ c }{Context}  &   \multicolumn{4}{ c }{Dosage (nM)}\\
                    & conditions& Drug & Dose                    &10 & 100 & 1000 & 10000    \\
    \midrule 
    \midrule
    Identity                     &     &             &            & $0.747_{0.126}$           & $0.654_{0.260}$           & $0.503_{0.332}$           & $0.227_{0.212}$           \\
    Monge-OOD           &  32  &            &            & $0.715_{0.165}$           & $0.238_{0.294}$           & $0.611_{0.288}$           & $0.663_{0.249}$           \\
    CMonge-DD-RDkit-OOD    &  32  & \checkmark & \checkmark & $0.466_{0.261}$           & $0.285_{0.265}$           & $0.287_{0.263}$           & $0.618_{0.316}$           \\
    CMonge-DD-MoA-OOD      &  32  & \checkmark & \checkmark & $\mathbf{0.879_{0.086}}$  & $\mathbf{0.887_{0.097}}$  & $\mathbf{0.894_{0.091}}$  & $\mathbf{0.908_{0.137}}$  \\
    chemCPA-DDCellLine-OOD &  32  & \checkmark & \checkmark & $0.655_{0.188}$           & $0.571_{0.239}$           & $0.429_{0.307}$           & $0.211_{0.191}$           \\
    \bottomrule
    \end{tabular}
\end{table}

\tabcolsep=0.10cm
\begin{table}[!htb]
\centering\footnotesize
\caption{Evaluation of conditional drug and dose experiments. Results are compared based on the Coefficient of Determination between the predicted and target feature means ($R^2$). Results show average and standard deviation across all conditions, evaluated in a leave-9-drugs-out cross-validation setting.}
\label{table:large_drug_dose_ood}
\begin{tabular}{lccccccc}
    \toprule
    Model           & conditions& \multicolumn{2}{ c }{Context}  &   \multicolumn{4}{ c }{Dosage (nM)}\\
                    & seen      & Drug & Dose                    &10 & 100 & 1000 & 10000    \\
    \midrule 
    \midrule
    CMonge-DrugDose-MoA-OOD      &  712 & \checkmark & \checkmark & $\mathbf{0.931_{0.042}}$           & $\mathbf{0.914_{0.046}}$           & $\mathbf{0.917_{0.053}}$           & $\mathbf{0.900_{0.059}}$           \\
    CMonge-DrugDose-RDkit-OOD    &  712 & \checkmark & \checkmark & $0.821_{0.173}$           & $0.815_{0.168}$           & $0.801_{0.182}$           & $0.781_{0.187}$           \\
    chemCPA-DrugDoseCellLine-OOD &  712 & \checkmark & \checkmark & $0.836_{0.154}$           & $0.815_{0.178}$           & $0.792_{0.205}$           & $0.760_{0.211}$           \\
    \bottomrule
    \end{tabular}
\end{table}

\tabcolsep=0.10cm
\begin{table}[!htb]
\centering\footnotesize
\caption{Evaluation of conditional drug and dose experiments. Results are compared based on the Wasserstein distance between predicted and measured cells. Results show average and standard deviation across all conditions, evaluated in a leave-9-drugs-out cross-validation setting.}
\label{table:large_drug_dose_ood_wasserstein}
\begin{tabular}{lccccccc}
    \toprule
    Model           & conditions& \multicolumn{2}{ c }{Context}  &   \multicolumn{4}{ c }{Dosage (nM)}\\
                    & seen      & Drug & Dose                    &10 & 100 & 1000 & 10000    \\
    \midrule 
    \midrule
    CMonge-DrugDose-MoA-OOD      &  712 & \checkmark & \checkmark & $\mathbf{3.864_{0.633}}$           & $\mathbf{3.873_{0.643}}$           & $\mathbf{3.878_{0.603}}$           & $\mathbf{3.825_{0.578}}$           \\
    CMonge-DrugDose-RDkit-OOD    &  712 & \checkmark & \checkmark & $4.205_{0.887}$           & $4.232_{0.896}$           & $4.245_{0.819}$           & $4.156_{0.747}$           \\
    chemCPA-DrugDoseCellLine-OOD &  712 & \checkmark & \checkmark & $4.707_{0.913}$           & $4.767_{0.976}$           & $4.812_{1.070}$           & $4.947_{1.271}$           \\
    \bottomrule
    \end{tabular}
\end{table}

\begin{table}[!htb]
\centering
\caption{Evaluation of drug effect perturbations, treated with 9 different drugs. Results are compared based on the Coefficient of Determination between the predicted and target feature means ($R^2$). The average and standard deviation are reported of the 9 experiments per model.}
\label{table:benchmark}
\begin{tabular}{lcccc}

    \toprule
    Model &\multicolumn{4}{ c }{Dosage (nM)}\\
    &10&100&1000&10000\\
    \midrule
    \midrule
    Monge & $0.947\pm0.024$ & $0.931\pm0.043$ & $0.951\pm0.022$ & $0.962\pm0.027$ \\
    scGen & $0.859\pm0.106$ & $0.839\pm0.134$ & $0.847\pm0.067$ & $0.886\pm0.079$ \\
    ICNN & $0.969\pm0.013$ & $0.955\pm0.034$ & $0.962\pm0.017$ & $0.972\pm0.012$ \\
    Identity & $0.666\pm0.177$ & $0.575\pm0.276$ & $0.426\pm0.288$ & $0.215\pm0.189$ \\
    \bottomrule
    \end{tabular}
\end{table}

\end{document}